\documentclass[sigconf]{acmart}

\makeatletter
\def\@ACM@checkaffil{
    \if@ACM@instpresent\else
    \ClassWarningNoLine{\@classname}{No institution present for an affiliation}%
    \fi
    \if@ACM@citypresent\else
    \ClassWarningNoLine{\@classname}{No city present for an affiliation}%
    \fi
    \if@ACM@countrypresent\else
        \ClassWarningNoLine{\@classname}{No country present for an affiliation}%
    \fi
}
\makeatother

\usepackage{graphicx}
\usepackage[linesnumbered,ruled]{algorithm2e}
\usepackage{listings}
\usepackage{xcolor}
\usepackage{wrapfig}
\usepackage{multirow}
\usepackage{balance}  
\usepackage{makecell} 
\usepackage{subcaption} 
\usepackage{amsmath} 
\usepackage{comment}
\usepackage{tcolorbox}
\usepackage{algorithmic}

\usepackage{amsfonts} 
\usepackage{float}

\AtBeginDocument{%
  }

\setcopyright{acmlicensed}
\copyrightyear{2026}
\acmYear{2026}
\acmDOI{XXXXXXX.XXXXXXX}
\acmConference[ACM SIGMOD '26]{the 2026 International Conference on Management of Data}{May 31 -- June 05, 2026}{Bengaluru, India}
\acmISBN{978-1-4503-XXXX-X/2026/04}

\begin{document}

\title{CONE: Embeddings for Complex Numerical Data Preserving Unit and Variable Semantics}

\author{Gyanendra Shrestha}
\affiliation{%
    \institution{Florida State University}}

\author{Anna Pyayt}
\affiliation{%
    \institution{University of South Florida}}
    
\author{Michael Gubanov}
\affiliation{%
    \institution{Florida State University}}
     
\begin{abstract}
Large pre-trained models (LMs) and Large Language Models (LLMs) are typically effective at capturing language semantics and contextual relationships. However, these models encounter challenges in maintaining optimal performance on tasks involving numbers. Blindly treating numerical or structured data as terms is inadequate -- their semantics must be well understood and encoded by the models. In this paper, we propose \texttt{CONE}, a hybrid transformer encoder pre-trained model that encodes numbers, ranges, and gaussians into an embedding vector space preserving distance. We introduce a novel composite embedding construction algorithm that integrates numerical values, ranges or gaussians together with their associated units and attribute names to precisely capture their intricate semantics. We conduct extensive experimental evaluation on large-scale datasets across diverse domains (web, medical, finance, and government) that justifies \texttt{CONE}'s strong numerical reasoning capabilities, achieving an F1 score of 87.28\% on DROP, a remarkable improvement of up to 9.37\% in F1 over state-of-the-art (SOTA) baselines, and outperforming major SOTA models with a significant Recall@10 gain of up to 25\%.
\end{abstract}





\maketitle

\section{Introduction}
\begin{figure}[t]
 \centering
  \includegraphics[width=1\columnwidth]{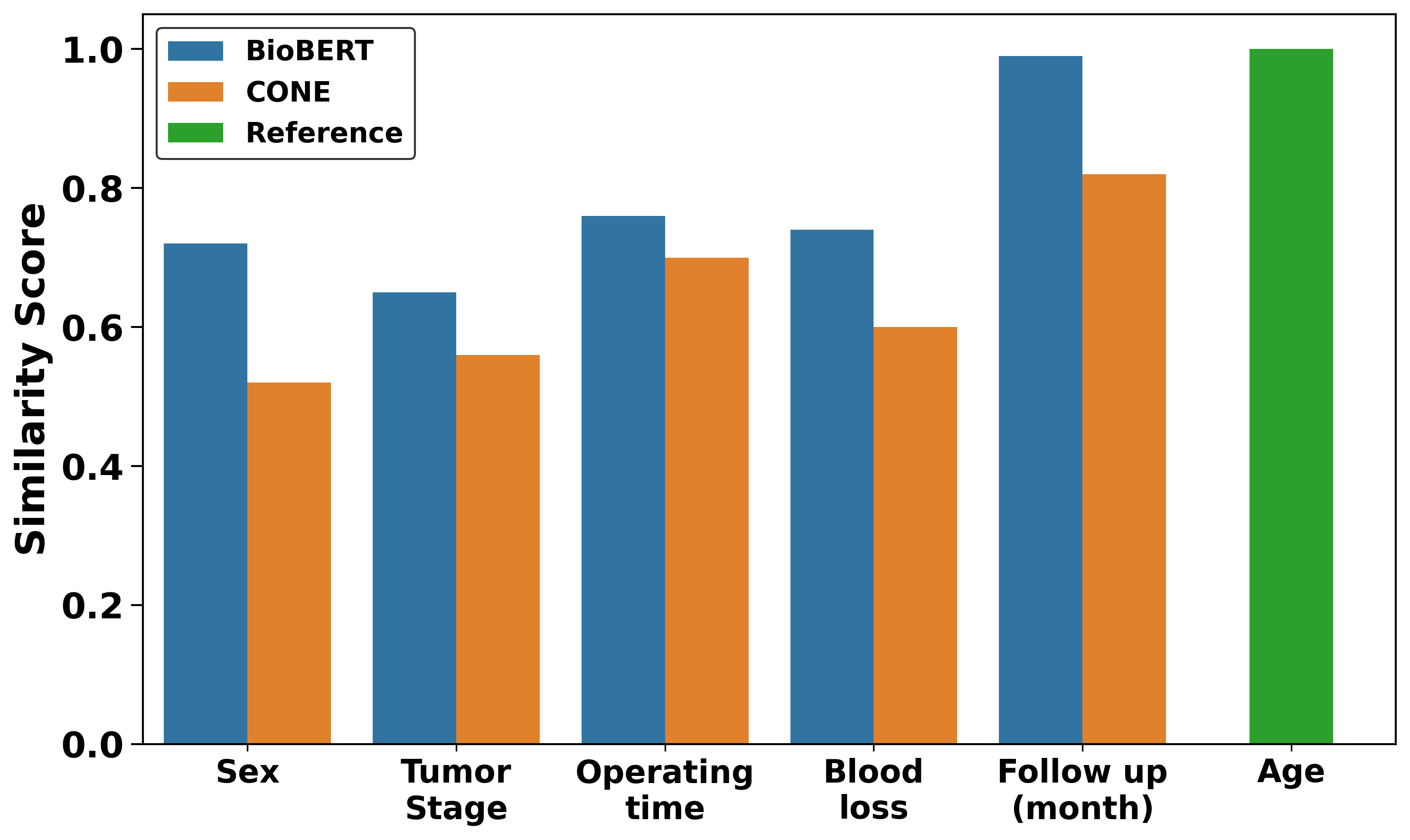}
   \captionsetup{skip=2pt} 
 \caption{Distance Between \textit{Age} and Other Attributes, Calculated Using BioBERT and \texttt{CoNE}.}
 \vspace{-2ex}
 \label{fig:angle1}
\end{figure}

Natural language text typically contains both numerical values and words. Numerical data is especially ubiquitous and important in Science, Technology, Engineering, and Mathematics (STEM) fields, where precise numerical representations are critical.  Numerical values can be expressed as digits (e.g., `15'), in alphabetic form (e.g., `fifteen') both accompanied with units (e.g., `15 miles'). Units play a major role in correct comprehension and differentiation of quantities (e.g., 10 miles versus 10 kilograms). Numerical values in structured data carry additional semantics, such as being a value for a specific attribute and can also be accompanied by units. Despite the impressive performance of recent language models in natural language understanding \cite{BERT}, they still struggle with tasks involving numerical and structured data. Multiple studies show that numerical embeddings from pre-trained LMs do not capture correctly the semantics of numerical values \cite{naik2019exploring, wallace2019nlp, BERT, elmo, zhang2020language, lee2020biobert, sundararaman2020methods, yarullin2023numerical, genbert, ran2019numnet}. One reason for this limitation stems from the tokenization strategies used in these models, which blindly treat numbers and words alike when deriving embedding vectors. For example, BERT \cite{BERT} uses a subword tokenization approach \cite{sennrich2016neural}, which can split a number just like a word into several parts. Consider the number 28,600: it might be tokenized as ``28” and ``–600” completely distorting the original semantics. It is also important to note that different tokenizers can produce different segmentations. Finally, all these models overlook the fundamental differences between words and numbers as described in measurement theory \cite{stevens1946theory}. Words are typically {\em nominal} (used for labeling without inherent order or magnitude), whereas numbers follow {\em interval} or {\em ratio} scales, where distance, order, and proportionality are crucial. Regular word embeddings capture the words' semantic similarity, but lack mechanisms to correctly encode numerical data as well as fail to preserve its fundamental numerical properties. The issue becomes magnified by structured data that unlike 1D text has 2D-context, where numbers are associated with specific attributes. For example, two columns \textit{Age} and \textit{Follow-up (month)} in Figure \ref{fig:num_table} contain similar distributions of numbers, yet they are semantically distinct and that should be reflected in their corresponding vectorized representations. Pre-trained LMs often struggle to differentiate between such attributes (Figure \ref{fig:angle1}), leading to poor performance in many numerical reasoning tasks.

\begin{table*} [t]
 \caption{\texttt{CONE} versus SOTA Models in Key Numerical Reasoning Capabilities and Support for Complex Numerical Data.}
 \vspace{-2ex}
  \label{tab:comparisonTable}
 \centering
 \small
  \begin{tabular}{|c|c|c|c|c|c|c|c|c|c|}
    \hline  
        \textbf{Features} & \textbf{BERT} & \textbf{ELMo} & \textbf{NumBERT} & \textbf{BioBERT} & \textbf{DICE} & \textbf{AeNER} & \textbf{GenBERT}  & \textbf{NumNet} & \textbf{\texttt{CONE}} \\
     \hline
     Numeration & limited & limited & yes & limited & yes & yes  & yes & yes & yes \\
     \hline  
     Magnitude & yes & yes & yes & yes & yes & yes & yes & yes & yes \\
     \hline  
     List maximum & limited & better than BERT & -  & limited & yes & yes & yes & yes & yes \\
     \hline  
     Decoding & limited & better than BERT & - & Limited & yes & yes & yes & yes & yes \\
     \hline  
     Addition & limited & limited & - & limited & yes & yes & yes & yes & yes \\
     \hline  
     Scalar Probing & some & limited & good & limited & - & - & - & - & yes \\
     \hline  
     Text & yes & yes & yes & yes & yes (as textual & yes & yes & yes & yes \\
         &     &      &     &     &  numbers) &       &      &    &      \\
     \hline  
     Tabular Numerical Data & no & no & no & no & no & yes & no & no & yes \\
  \hline
\end{tabular}

\end{table*}

\begin{figure}[b]
\centering
\setlength{\tabcolsep}{4.5pt}
\renewcommand{\arraystretch}{1.3}
\scriptsize
\begin{tabular}{|c|c|c|c|c|c|c|c|}
\hline

\textbf{Age} &
\textbf{Sex} &
\makecell{\textbf{Tumor}\\\textbf{Stage}} &
\makecell{\textbf{Operating }\\\textbf{time}} &
\makecell{\textbf{Blood}\\\textbf{loss}} &
\makecell{\textbf{Follow-up}\\\textbf{(months)}} &
\makecell{\textbf{BP}\\\textbf{(mmHg)}} &
\makecell{\textbf{BMI}\\\textbf{(kg/$\text{m}^2$)}}\\
\hline
 28 & F & S1--3 & 7 hrs & 3,000 mL & 30 & 76-118 & 21.8 $\pm$ 2.9 \\
\hline
 34 & F & S1--3 & 580 min & 5 L & 32 & 80-122 & 23.4 $\pm$ 3.2\\
\hline
 36 & F & S1--3 & 6 hrs & 3.5 L & 28 & 82-125 & 22.5 $\pm$ 3.17\\
\hline
 42 & M & S1--2 & 500 min & 3,000 mL & 25 & 85-130 & 24.1 $\pm$ 3.5\\
\hline
33 & F & S1--3 & 400 min & 3.2 L & 20 & 78-120 & 22.2 $\pm$ 3.0\\
\hline
 31 & M & S1--2 & 500 min & 3,000 mL & 15 & 84-128 & 23.6 $\pm$ 3.4\\
\hline
\end{tabular}
\vspace{0.5ex}
\caption{A Table with Numerical Data Including Ranges and Gaussians Accompanied by Units.}
\label{fig:num_table}
\end{figure}

Recently, numerical embeddings started gaining more attention \cite{zhang2020language, sundararaman2020methods, yarullin2023numerical, genbert, ran2019numnet}. Table \ref{tab:comparisonTable} compares \texttt{CONE} with SOTA baselines. The comparison is based on reported results in the original papers and prior literature. 
None of these approaches fully capture semantics of numerical data having units, ranges, and gaussians. This fundamental limitation hinders their ability to fully comprehend and correctly encode complex numerical data found in text and structured data, where their semantics depend on the associated attributes and units (e.g., \textit{Age: 50 years} vs. \textit{Weight: 50 kg}).

To address these limitations, we describe \texttt{CONE}, a novel model for constructing composite embeddings that correctly encode and encapsulate semantics of complex numerical data including ranges and gaussians with units. \texttt{CONE} is built on top of pre-trained LMs, such as BERT \cite{BERT} and BioBERT \cite{lee2020biobert} extending their capabilities to handle complex numerical semantics. By encoding numerical values within their complex structural context, this {\em composite} embedding approach ensures that numbers with differing units and/or attribute/variable, such as \textit{5 km} and \textit{5 kg} or \textit{5 mg:dosage} and \textit{5 mg:weight}, are not being blindly treated alike, even though their numerical value is identical (5). On top of that, \texttt{CONE} comprehends and encodes correctly semantics of numerical ranges (e.g., \textit{[5-10 years]:Age}) and gaussians (e.g., \textit{[1302±0.25 nm, mean $\pm$ SD]:Particle Size}) correctly allowing the model not to lose their real semantics. Through extensive experiments on popular downstream tasks with complex structured data we demonstrate that \texttt{CONE} significantly outperforms the widely used SOTA baselines, such as TAPAS \cite{herzig2020tapas}, NumNet \cite{ran2019numnet}, NC-BERT \cite{kim2022exploiting}, Magneto \cite{liu2025magneto} and \textit{general-purpose embedding models} \cite{chen2024m3, zhang2024jasper, zhang2025qwen3, hu2025kalmembedding}. By incorporating unit-aware and attribute-specific embeddings, our model ensures that complex numerical data is represented correctly, improving both retrieval and predictive downstream tasks.

Our key contributions are as follows:
\begin{itemize}
    \item Novel composite embedding structure for complex numerical data that incorporates the numerical value, unit, and attribute, allowing the model to accurately encode and comprehend numbers with their complex associated contexts. This ensures that numbers with different units and attributes are distinct and the encoding preserves their fundamental numerical properties.
    \item Special embeddings for numerical ranges and gaussians preserving their semantics.
    \item Two novel algorithms for the embedding vector components pre-computation (Algorithm \ref{alg:CONE-train}) and complete embedding vector composition (Algorithm \ref{alg:composite}).
    \item Extensive experimental evaluation of \texttt{CONE} on large-scale datasets from different domains, demonstrating its superior performance on popular downstream tasks.
\end{itemize}

The rest of the paper is structured as follows. We begin with preliminaries, then describe our methodology, followed by rigorous embedding distance analysis, extensive experimental evaluation, related work, and conclusion.

\section{Preliminaries}
\label{sec:prelim}

While language models demonstrate impressive performance on capturing textual semantics, they remain limited in their ability to capture the semantics of numerical values or complex structured data, where numbers inherit their semantics from the \textit{attribute} (e.g., \textit{Age}, \textit{Dose}, \textit{BMI}), the \textit{units} in which they are expressed (\textit{years}, \textit{mg}, \textit{kg/$\text{m}^2$}),
and the \textit{structure} of the numerical values themselves (single
value, ranges, or gaussians). A numerical value, such as ``30'' may correspond to \textit{30 years of age}, \textit{30 months of follow-up}, or \textit{30\,mm of tumor size}; its meaning depends entirely on its associated attribute and unit. Standard language models (e.g., BioBERT) do not encode these distinctions correctly, hence may confuse  \textit{attributes} that are numerically identical or lexically similar, but semantically different.

Let a table be $T = (C, R)$, where
$C = \{c_1,\dots,c_m\}$ is the set of $m$ columns with attribute names
$A = \{a_1,\dots,a_m\}$,
and $R = \{r_1,\dots,r_n\}$ is the set of $n$ tuples with 
$r_i = [v_{i1},\dots,v_{im}]$ the corresponding cell values.

\textsc{Definition 1 (Column Similarity).}
Two columns $c_1$ and $c_2$ are similar iff they have similar attribute values \cite{hasan2024dominance}.
We describe the column similarity measure more precisely below.

Each column $c_j$ contains attribute values
$\{v_{1j},\dots,v_{nj}\}$. For each $v_{ij}$, we compute a composite embedding $E^{\text{comp}}(a_j,v_{ij},u)$ (Algorithm \ref{alg:composite}) that encodes the attribute name $a_j$, $v_{ij}$, and unit $u$.
The embedding vector of column $c_j$ is then obtained by aggregation:
\begin{equation}
\footnotesize
{E}(c_j) = \sum_{i=1}^{n} E^{\text{comp}}(a_j, v_{ij}, u)
\end{equation}

Similarity between two columns $c_1$ and $c_2$ is computed by calculating the cosine of the angle between their aggregated embedding vectors \cite{hasan2024dominance}:
\begin{equation}
\footnotesize
Sim_{{col}}(c_1, c_2)
= \cos\big(E(c_1), E(c_2)\big)
\end{equation}

\textsc{Definition 2 (Tuple Similarity).}
Two tuples $r_1$ and $r_2$ are similar iff they have similar cell values. We describe the tuple similarity measure more precisely below.

Each tuple $r_i = [v_{i1},\dots,v_{im}]$ consists of cell values $v_{ij}$ corresponding to each attribute $ a_j\in A$. The embedding of tuple $r_i$ is obtained by aggregation:
\vspace{-2ex}
\begin{equation}
\footnotesize
{E}(r_i) = \sum_{j=1}^{m} E^{\text{comp}}(a_j, v_{ij}, u)
\end{equation}

Similarity between two tuples $r_1$ and $r_2$ is computed by calculating the the cosine of the angle between their aggregated embedding vectors:
\begin{equation}
\footnotesize
Sim_{{tuple}}(r_1, r_2)
= \cos\big(E(r_1), E(r_2)\big)
\end{equation}


\textsc{Example 1.} \textit{Consider a table in Figure \ref{fig:num_table}. `Age (years)' can be stored   as \{28, 34, 36, 42, 33, 31\}, \{29, 31, 35, 42, 32, 32\}, or as a range such as
``30--45''. Despite different representation, these belong to the same
demographic concept and must therefore be embedded close to each other in the vector space. In
contrast, attributes such as `Follow-up Duration (months)' with values
\{20, 25, 30, 36, 40, 32\} or \textit{Tumor Size (mm)} with values \{30, 35, 42, 22, 26, 32\} may
numerically overlap with `Age', but encode completely different variables. Hence, they should not be positioned close to `Age' in
the embedding vector space. The only attributes that should exhibit high similarity
to `Age' in Figure \ref{fig:angle1} are other attributes that may contain some `Age' values. The attributes, such as `Follow-up', or `Operating Time' must remain well separated, regardless of any numeric overlap due to their  different semantics.}

We can see in Figure \ref{fig:angle1} that BioBERT cannot distinguish  \textit{Age} and \textit{Follow-up(month)} originating from the table in Figure \ref{fig:num_table}. Their  cosine similarity score is $0.9998$, hence the angular distance is very small - $1.02^\circ$, which means that these two vectors constructed by BioBERT nearly overlap with each other. 
This is due to very similar distributions of numerical values from these two columns, and "\textit{blindness}" of vanilla BERT embedding vectors to differences in units and attributes (i.e. lack of numerical and unit semantics). BioBERT’s high similarity scores are driven primarily by textual similarity between \textit{attribute names} (e.g., \textit{month} appearing semantically similar to \textit{Age}) rather than by the more intricate semantics of the numerical values. Not only BERT and BioBERT, but also other SOTA \textit{general-purpose embedding models} \cite{chen2024m3, zhang2024jasper, zhang2025qwen3, hu2025kalmembedding} do not perform well on tasks involving numbers, also because they treat numbers as regular terms.

In large-scale datasets, this issue is exacerbated by attribute heterogeneity and data representation variety. The tables come from a myriad of  different sources, which use different naming conventions, schemata, formats. Further, they use different synonyms, abbreviations in tables' headers, which complicates the problem. There will also be variance in both textual and numerical data formats for the same attributes, which poses additional challenges constructing the embeddings correctly mimicking the data semantics. To overcome these challenges we propose a more complex -- {\em composite} structure for the embedding vectors and a modified encoder architecture that accurately encode semantics of complex numerical data. 

\textsc{Definition 3 (Serialization).}
At a column-level, we serialize each column $c_j$ independently as: $\mathcal{S}_{\text{col}}(c_j) \mathrel{\mathop{:}\!\text{-}} [\text{CLS}] \, a_j \, [\text{SEP}]\, v_{1j}\, [\text{SEP}]\, \\ \dots\, v_{nj}\, [\text{SEP}]$. I.e., we place the column name first, followed by all cell values, using \text{[SEP]} tokens to delimit values and a single \text{[CLS]} token at the start (where [CLS] and [SEP] follow the BERT tokenization standard \cite{BERT}).
At a tuple-level, we serialize a row $r_i$ as
$\mathcal{S}_{\text{row}}(r_i)\mathrel{\mathop{:}\!\text{-}}[\text{CLS}] \, a_1 \, v_{i1} \, [\text{SEP}] \, \dots a_m \, v_{im} \, [\text{SEP}]$. I.e., each attribute name is interleaved with its corresponding cell value and attribute--value pairs are separated by \text{[SEP]} tokens.

We experimented with multiple alternative serialization strategies, which proved less effective. For consistency, we use the same serialization for BioBERT and other models in our experiments. Throughout the paper, we use the term \textit{attribute} to refer to a column header (i.e., the column name), whereas \textit{column} denotes the entire column, including both the header and its cell values.

\textsc{Example 2}. \textit{Consider the tuples and columns in the table shown in Figure \ref{fig:num_table}. We serialize the column with attribute `Age' in the table as: [CLS]\, \text{Age}\, [SEP]\, 28\, [SEP]\, 34\, [SEP]\, 36\, [SEP]\, 42\, [SEP]\, 33\, [SEP]\, 31\, [SEP]. Similarly, we serialize the tuple (the first row) in the table as: [CLS]\, Age 28\, [SEP]\, Sex F \, [SEP]\, Tumor Stage S1-3\, [SEP]\, Operating time 7 hrs\, [SEP]\, Blood loss 3000 mL\, [SEP]\, Follow-up (month) 30\, [SEP]\, BP (mmHg) 76-118\, [SEP]\,BMI (kg/$\text{m}^2$) 21.8 $\pm$ 2.9\, [SEP].}

We modified BERT architecture \cite{BERT, lee2020biobert} to accommodate advanced semantics of complex numerical data having units, numbers, ranges, and gaussians. Each Transformer layer includes a multihead self-attention sub-layer, where each token attends to all the tokens. Let the layer input ${H} = [h_1, h_2, \dots, h_n]^T \in \mathbb{R}^{n \times d}$
correspond to sequence \( S \), where \( d \) is the hidden dimension, and \( h_i \in \mathbb{R}^{d \times 1} \) is the hidden representation at position \( i \). For a single-head self-attention sub-layer, the input $H$ is projected by three matrices ${W}^Q \in \mathbb{R}^{d \times d_K}$, ${W}^K \in \mathbb{R}^{d \times d_K}$, and ${W}^V \in \mathbb{R}^{d \times d_V}$ to the corresponding representations $Q$, $K$, and $V$:
\begin{equation}
   Q = H {W}^Q, \quad V = {H} {W}^V, \quad {K} = {H} {W}^K 
\end{equation}

Then, the output of this single-head self-attention sub-layer is calculated as:
\vspace{-2ex}
\begin{equation}
  \text{Attn}({H}) = \text{softmax}\left(\frac{{Q} {K}^T}{\sqrt{d_K}}\right){V}  
\end{equation}

As illustrated in Figure \ref{fig:angle1}, our method reduces the similarity between \textit{Age} and \textit{Follow-up} to 0.82. When ranking column similarity to \textit{Age}, our method places all Age-related columns at the top, while \textit{Follow-up} does not appear among the top candidates. Our method is explicitly designed to encode such distinctions by jointly capturing \emph{attribute}, \emph{unit}, and \emph{numerical values}, ensuring that semantically equivalent numerical attributes are embedded closely while semantically different ones remain well separated. Likewise another source of ambiguity arises when attributes share the same name, but use incompatible units. For instance, the attribute \textit{Dose} may refer to 40--60\,mg (drug mass), 4--6\,mL (drug volume), or 1.0--1.5\,IU/kg (biological potency).
Text-based embeddings consider these nearly the same, because the term `\textit{Dose}' dominates their embedding representation and the unit mismatch is not considered. Our method incorporates unit semantics explicitly, ensuring that \textit{Dose (mg)}, \textit{Dose (mL)}, and \textit{Dose (IU/kg)}
are embedded as distinct concepts. Our embedding representation is general-purpose and does not require any modification for different downstream tasks.

\section{\texttt{CONE} Model}\label{sec:model}
\begin{figure}[htpb]
 \centering
   \includegraphics[width=1\columnwidth] 
   {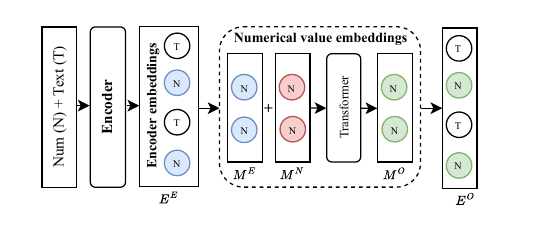}
   \captionsetup{skip=2pt} 
 \caption{\texttt{CONE} Architecture including the Encoder and Numerical Value Embeddings. The Encoder Embeddings for Numbers ($M^E$, shown in blue) are Fused with Numerical Value Embeddings ($M^N$, pink) and Refined Through a Transformer to Produce the Contextualized Representations ($M^O$, green).}
 \vspace{-2ex}  
 \label{fig:architecture}
\end{figure}

Our model architecture is illustrated in Figure \ref{fig:architecture}. It enhances standard transformer encoders with numerically-aware embeddings, improving the model's ability to understand and reason about numerical values.

\subsection{Encoder}
We adopt a transformer-based language model as the encoder \cite{BERT, yarullin2023numerical}, initialized with BioBERT \cite{lee2020biobert} weights and fine-tuned on our datasets. In this work, we focus on numerical columns/tuples in tables. The input to the encoder is a tokenized concatenation of [CLS], and table's tuple or column, where the table is flattened along row/column and the cells are separated by the [SEP] token using a serialization function $\mathcal{S}$ (see Definition~3). Given a tokenized input sequence of length $l$, the encoder generates a corresponding contextualized representation denoted as:
\begin{equation}
{E}^E = [e_1, e_2, \dots, e_l] \in \mathbb{R}^{l \times d}
\end{equation}
where $d$ represents the embedding dimension (we use $d = 768$ throughout).

We do not modify the default tokenizer of the underlying encoder for textual tokens, which therefore follow the standard tokenization process \cite{lee2020biobert}. However, we override the tokenizer’s behavior for numberical data (see Section \ref{sec:nm}): each number is treated as a single token, rather than being split into multiple subword tokens as in BERT/BioBERT, whenever no corresponding embedding exists. The embedding for each numerical token is learned through our custom architecture in Figure \ref{fig:architecture}.

\subsection{Embeddings for Numerical Values} \label{sec:nm}
We propose \textit{numerical fusion} embeddings that jointly capture symbolic context and quantitative magnitude. This is achieved by augmenting the standard token embeddings of numerals with numerical value embeddings \cite{sundararaman2020methods,jin2021numgpt, berg2020empirical, spithourakis2018numeracy, gorishniy2022embeddings, thawani2021numeracy} and integrating them through attention \cite{yarullin2023numerical}. Although we focus on encoder-based models in this work, the \texttt{CONE} framework is not architecture-specific. The numerical attention fusion mechanism can be applied to both encoder and decoder architectures by replacing their token embeddings with the fused representation, followed by the lightweight transformer block for number-specific reasoning.

We extract numeric tokens (i.e. numerals) from the input (e.g. column or row) and use DICE \cite{sundararaman2020methods} to generate numerical value embeddings $M^N$, one component of our embedding vector that enhances the representation of numerical properties (e.g., magnitude). One can also use other existing methods \cite{jin2021numgpt, berg2020empirical, spithourakis2018numeracy, gorishniy2022embeddings, thawani2021numeracy}.

Let ${I}^{\text{num}} = \{i_1, i_2, \dots, i_s\}$ be the set of indices of numeric tokens. Each index corresponds to the relative position of a numeral token in the input sequence. For each $i_j \in I$, we retrieve the corresponding encoder embedding from $E^E$ and denote the subset of these embeddings ${M}^E \in \mathbb{R}^{s \times d}$ as follows: 
\begin{equation} \label{test1}
M^{E}_j = e_{i_j}, \quad \text{for } j = 1, \ldots, s
\end{equation}

We compute a \textit{fused} representation by element-wise summation of the contextual embeddings ${M}^E$ and the numerical value embeddings ${M}^N$, and feed the result into a transformer block to enable number-specific reasoning. The resulting output ${M}^O$, which provides contextualized embeddings for each numeral token, is then used to update the corresponding embeddings in the encoder output ${E}^E$. This yields the final embedding ${E}^O$, which incorporates both contextual and numerical semantics.

\begin{figure*}[htpb]
 \centering
   \includegraphics[width=1\linewidth] {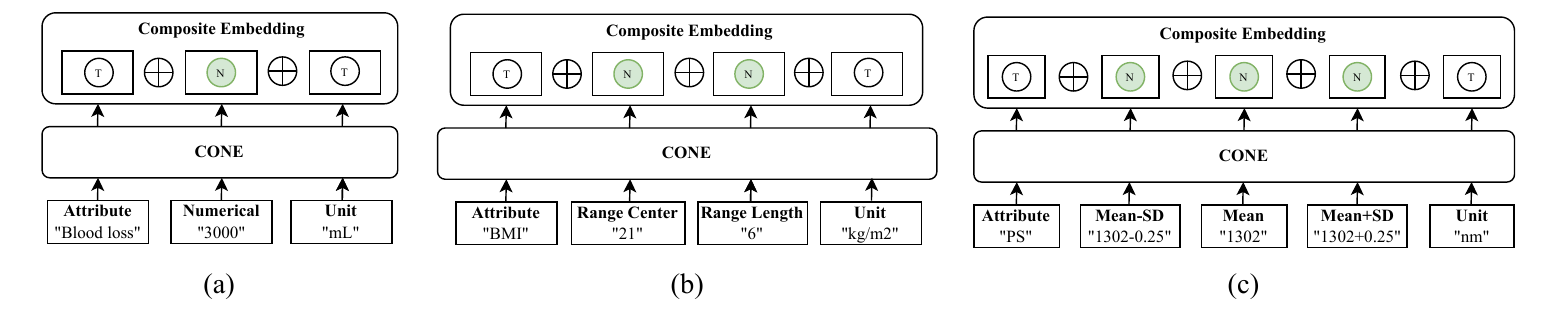}
   \captionsetup{skip=2pt} 
 \caption{Composite Embedding Structure for (a) Complex Numerical Data, (b) Numerical Ranges, and (c) Gaussians. T: Text, N: Numerical Value, SD: Standard Deviation.}
    \vspace{-2ex}  
 \label{fig:comp}
\end{figure*}

\begin{algorithm}[htpb]
\caption{\texttt{CONE} Training via Masked Numeral Prediction}
\label{alg:CONE-train}
\begin{algorithmic}[1]
\STATE \textbf{Input:} Tokenized input $x = [x_1, \dots, x_l]$; encoder $f_\theta$ (initialized with BioBERT); numerical value embedding $\text{NumEmbed}(\cdot)$; transformer block $g_\phi$; masking ratio $p$; Adam optimizer $\mathcal{O}$
\STATE \textbf{Output:} Trained parameters $\theta$ and $\phi$
\STATE $I^{\text{num}} \gets \{ i \in [1..l] \mid x_i \text{ is a numeral} \}$ \hfill // indices of numeral tokens
\STATE $\tilde{x} \gets \text{MaskNumerals}(x, I^{\text{num}}, p)$ \hfill // replace subset with [MASK]
\STATE $E^E \gets f_\theta(\tilde{x}) \in \mathbb{R}^{l \times d}$ \hfill // contextual embeddings
\STATE $M^E \gets [E^E_i]_{i \in I^{\text{num}}} \in \mathbb{R}^{s \times d}$ \hfill // $s = |I^{\text{num}}|$
\STATE $M^N \gets [\text{NumEmbed}(x_i)]_{i \in I^{\text{num}}} \in \mathbb{R}^{s \times d}$
\STATE $M^{F} \gets M^{E} + M^{N}$ \hfill // fusion by element-wise sum
\STATE $M^{O} \gets g_\phi(M^{F})$ \hfill // number-specific reasoning
\STATE $(\hat{y}, \hat{c}) \gets \text{Heads}(M^{O})$ \hfill // magnitude $\hat{y}$ and class $\hat{c}$
\STATE $(y, c) \gets \text{GroundTruthMagnitudesAndClasses}(x, I^{\text{num}})$
\STATE $\mathcal{L}_{\text{mag}} \gets \sum_{j=1}^{s} (\log y_j - \log \hat{y}_j)^2$
\STATE $\mathcal{L}_{\text{cls}} \gets \sum_{j=1}^{s} \text{CEL}(c_j, \hat{c}_j)$
\STATE $\mathcal{L}_{\text{num}} \gets \mathcal{L}_{\text{mag}} + \mathcal{L}_{\text{cls}}$
\STATE $\theta, \phi \gets \mathcal{O}(\theta, \phi, \nabla_{\theta,\phi}\mathcal{L}_{\text{num}})$
\STATE \textbf{return} Updated parameters $\theta$ and $\phi$
\end{algorithmic}
\end{algorithm}

We train the model using a \textit{masked numeral prediction} task \cite{spithourakis2016numerically, lin2020birds, sakamoto2021predicting}, as outlined in Algorithm~\ref{alg:CONE-train}, where the model is trained to predict a masked numeral in the input. The objective function $\mathcal{L}_{\text{num}}$ (Equation~\ref{loss}) combines magnitude regression with classification loss:
\begin{equation} \label{loss}
\mathcal{L}_{\text{num}} = \sum_{i=1}^{N} \left\{ \left( \log(y_i) - \log(\hat{y}_i) \right)^2 + \text{CEL}_i \right\}
\end{equation}

\noindent where, $y_i$ denotes the ground-truth numerical magnitude, $\hat{y}_i$ is the magnitude of the numeral predicted by the model, $N$ is the number of masked numerals in the training, and $\text{CEL}_i$ is the cross-entropy loss for the $i$-th numeral token. After training, the numeral prediction head is discarded, and the final contextualized embedding $E^O$ is used for various downstream tasks. $E^O$ encodes numerals both in terms of their semantic role (context) and their quantitative value on the number line (magnitude).

\subsection{Composite Embedding Structure} \label{sec:composite_embed}
Prior studies \cite{yin2016learning,xu2018lifelong,bollegala2022learning,liu2015topical, liu2019single, wang2021automated,elmo} show that concatenated embeddings outperform single embeddings on tasks such as similarity and classification. Importantly, concatenation (CONC) retains each embedding in its original subspace, preserving source information without forcing it into a single space. This yields representations that are both expressive, by integrating complementary signals, and discriminative (e.g., \textit{Age = 50 years} vs. \textit{Weight = 50 kg}, where the identical value “50” is distinguished by attribute and unit context). Motivated by this, we propose a composite embedding structure that concatenates ($\oplus$) embeddings $E^O$ (see Figure~\ref{fig:architecture}) for the attribute, its value (scalar, range, or Gaussian), and the unit, thereby preserving the complementary contributions of each component to structured numerical data. Our composite embedding corresponds to combination of subspaces \cite{halmos2017finite}, ensuring that each component (attribute, value, unit) contributes independently to the overall distance (see Section \ref{sec:analysis} for a detailed distance analysis). For instance, when the attribute and unit are fixed, distances between composites reduce to differences in the value, so vector distances naturally reflect numeric proximity; the same property holds when either the attribute or the unit is fixed.

Figure \ref{fig:comp}(a) illustrates this process for a column \textit{Blood loss} from the table in Figure \ref{fig:num_table}, attribute \textit{Blood loss} has value “\textit{3,000}” in the first row, and the unit ``\textit{mL}” is specified along with the value. This composite embedding structure preserves the actual semantics of the numerical value. We call \textit{Range} (\textit{a-b}) an interval that includes all $x$ between two boundaries a and b, such that $a \leq x \leq b$. To construct the composite embedding for numerical \textit{Range} values (e.g., \textit{Age 1--5 years}), we represent each range using its center $\tfrac{a+b}{2}$ and length $|b-a|$, which provide an orthogonal representation of the range’s location and scale. The composite embedding is then obtained by concatenating the embeddings for the \textit{attribute}, \textit{center}, \textit{length}, and \textit{unit}. In Figure \ref{fig:comp}(b) we show this composite structure for \textit{BMI} (i.e. \textit{Body Mass Index}) “\textit{18-24}” \textit{kg/$\text{m}^2$}. For brevity, we call \textit{Gaussian} - a Gaussian distribution $mean\pm SD$ with $mean$ and SD, where SD denotes the Standard Deviation. The composite embedding for \textit{Gaussian} values has similar structure to that of \textit{Range}, where we concatenate the embeddings for the \textit{attribute}, ``$mean-SD$”, ``$mean$”, ``$mean+SD$”, and \textit{unit}. Figure \ref{fig:comp}(c) illustrates this structure using the example attribute \textit{PS (nm)} (\textit{Particle Size}) with the value ``$1302\pm0.25$”.

CONC embeddings have dimensionality proportional to the number of components (i.e., $k \cdot d$ for $k$ components). To avoid high and variable dimensionality in CONC and obtain a fixed-size representation, we adopt a slot-based concatenation scheme with zero padding and masking. We reserve a fixed set of slots $\mathcal{S}$ ($|\mathcal{S}|=5$ in our case), each of size $d$. Missing components are zero padded, and the corresponding binary mask bit is set to $m_s=0$.  Concatenating all slots yields $E \in \mathbb{R}^{|\mathcal{S}|d}$, which is then projected to a vector space with dimension $d_C=768$ using a linear autoencoder \cite{baldi1989neural,baldi2012autoencoders}. The reconstruction is $\hat E = W^\top W E$, where $W \in \mathbb{R}^{d_C \times (|\mathcal{S}|d)}$ denotes the encoder projection matrix and $W^\top$ the tied decoder, and training minimizes the masked reconstruction loss \cite{vincent2008extracting,he2022masked}.
\begin{equation}
    \mathcal{L}_{\text{embed}} = \| M \odot (\hat E - E) \|_2^2
\end{equation}
where $M \in \{0,1\}^{|S|d}$ is the broadcast mask derived from $\{m_s\}_{s\in \mathcal{S}}$, and $\odot$ denotes the Hadamard (element-wise) product. The encoded representation $z = W E$ is layer-normalized to produce the final composite embedding:
\begin{equation}
    E^{\text{comp}} = \mathrm{LayerNorm}(z)
\end{equation}

\begin{algorithm}[htbp]
\caption{Composite Embedding Construction}
\label{alg:composite}
\begin{algorithmic}[1]
\STATE \textbf{Input:} Attribute $a$, numeric value $v$ (scalar, range $[a,b]$, or Gaussian $(\mu,\sigma)$), unit $u$, slot size $d$, total slots $|S|$, projection matrix $W$
\STATE \textbf{Output:} Composite embedding $E^{\text{comp}}$
\STATE \textbf{if} $a$ exists \textbf{then} $E_a \gets \text{CONE}(a)$ \textbf{else} $E_a \gets \text{ZeroPad}(d)$ \textbf{end if}
\STATE \textbf{if} $v$ is scalar \textbf{then} $E_v \gets \text{CONE}(v)$
\STATE \textbf{else if} $v$ is range $(a,b)$ \textbf{then} \\
$E_v \gets \text{Concat}(\text{CONE}(\tfrac{a+b}{2}), \text{CONE}(|b-a|))$
\STATE \textbf{else if} $v$ is Gaussian $(\mu,\sigma)$ \textbf{then} \\
$E_v \gets \text{Concat}(\text{CONE}(\mu-\sigma), \text{CONE}(\mu), \text{CONE}(\mu+\sigma))$
\STATE \textbf{else} $E_v \gets \text{ZeroPad}(d)$ \textbf{end if}
\STATE \textbf{if} $u$ exists \textbf{then} $E_u \gets \text{CONE}(u)$ \textbf{else} $E_u \gets \text{ZeroPad}(d)$ \textbf{end if}
\STATE Concatenate embeddings: $E \gets [E_a, E_v, E_u]$ into $|S|$ slots of size $d$
\STATE Zero-pad unused slots and build binary mask $M$
\STATE Project: $z \gets W E$ \hfill // linear autoencoder projection
\STATE Reconstruct: $\hat{E} \gets W^{\top} z$
\STATE Compute masked reconstruction loss: $L_{\text{embed}} \gets \|M \odot (E - \hat{E})\|^2_2$
\vspace{-8pt}
\STATE Normalize: $E^{\text{comp}} \gets \text{LayerNorm}(z)$
\STATE \textbf{return} $E^{\text{comp}}$
\end{algorithmic}
\end{algorithm}

In our implementation, attribute names $a$, numerical values $v$, and unit symbols $u$ are extracted from tabular data for each row or column using a rule-based and regular-expression parser adapted from \cite{ptype-semantics}. The method in \cite{ptype-semantics} also supports unit canonicalization, which normalizes all unit variants into a consistent canonical form. The resulting triplet $(a,v,u)$ is then used to construct the composite embedding as described in Algorithm \ref{alg:composite}.

\section{Datasets} \label{sec:datasets}
We use the following five publicly available large-scale datasets in our experiments and we used the methods from \cite{kandibedala2025scalable} for complex metadata identification, which involves detecting multi-row, hierarchical headers and Bi-dimensional headers (i.e., columns and rows with metadata). This facilitates the construction of large-scale training and test sets for our experiments (Section \ref{subsec:cmtm}). 

\textbf{CancerKG} \cite{shresthatabular} dataset has 3.5M tables, extracted from all recent medical publications (up to year 12/2024) on different cancers, obtained via PubMed.com. The tables have 19.75M columns and 18.44M tuples total. The column/tuple values contain strings, numbers with and without units, ranges. \textbf{CovidKG} \cite{shresthatabular} is extracted from medical publications (up to year 2022) on COVID-19, obtained via PubMed.com. It contains 0.82M papers and 1.4M tables. \textbf{Webtables} \cite{webtable} contains around 1.65M tables extracted from Wikipedia pages. The corpus contains a large amount of factual knowledge on various topics ranging from sport events (e.g., Olympics) to artistic works (e.g., TV series). \textbf{CIUS} \cite{SAUS_CIUS} dataset is from the Crime In the US (CIUS) database and consists of 1,050 tables with information about crime offenses. \textbf{SAUS} \cite{SAUS_CIUS}, the 2010 Statistical Abstract of the United States (SAUS) from the U.S. Census Bureau comprises 1,368 tables covering topics like finance, business, crime, agriculture, and health. 

We also evaluate our model on \textbf{DROP} dataset \cite{dua2019drop}, a question answering (QA) benchmark that tests numerical reasoning skills such as counting, sorting, and addition. It contains 77,409 training, 9,536 development, and 9,622 test question–answer pairs. In addition, we use six datasets GDC, Magellan, WikiData, Open Data, ChEMBL and TPC-DI from \cite{liu2025magneto, koutras2021valentine} to evaluate and compare with recent schema matching (SM) solutions \cite{liu2025magneto, du2024situ}. Due to space constraints we refer the reader to \cite{liu2025magneto, koutras2021valentine} for statistics on these datasets.

\section{Analysis of Distance}\label{sec:analysis}
\begin{table}[t]
\centering
\caption{
Correlation of Analytic Distances (Section~\ref{sec:distance}) with Embedding Cosine Distances via Pearson’s $r$ and Spearman’s $\rho$.
}
\vspace{-2ex}
\label{tab:dis_eval}
 \setlength{\tabcolsep}{3pt}
\footnotesize
\begin{tabular}{lcc|cc|cc|cc}
\toprule
& \multicolumn{2}{c}{Numbers} & \multicolumn{4}{c}{Ranges} & \multicolumn{2}{c}{Gaussians} \\
\cmidrule(lr){2-3} \cmidrule(lr){4-7} \cmidrule(lr){8-9}
& \multicolumn{2}{c}{Abs. Diff.} & \multicolumn{2}{c}{Euclidean} & \multicolumn{2}{c}{Jaccard IoU} & \multicolumn{2}{c}{Wasserstein} \\
\cmidrule(lr){2-3} \cmidrule(lr){4-5} \cmidrule(lr){6-7} \cmidrule(lr){8-9}
Method & $r$ & $\rho$ & $r$ & $\rho$ & $r$ & $\rho$ & $r$ & $\rho$ \\
\midrule
BioBERT   & 0.067  & 0.064 & 0.398 & 0.355 & 0.267 & 0.248 & 0.038 & 0.039 \\
\texttt{CONE}   & \textbf{0.989} & \textbf{0.798} & \textbf{0.997} & \textbf{0.786} & \textbf{0.498} & \textbf{0.465} & \textbf{0.689} & \textbf{0.663} \\
\bottomrule
\end{tabular}
\end{table}

To validate  our composite embeddings for complex numerical data we check that each of its three components {\em separately} is contributing to the  distance as expected. Hence, we separately evaluated the effectiveness of fine-tuned embeddings for the number values, units, and attributes.

The Euclidean distance \cite{euclidean} between two \texttt{CONE} embedding vectors for numbers $\mathbf{x} = [x_1, y_2, \ldots, x_d]$ and $\mathbf{y} = [y_1, y_2, \ldots, y_d]$ is:
\begin{equation}
D_{xy} = \sqrt{\sum_{i=1}^d (x_i - y_i)^2}
\end{equation}




\subsection{Analytic distance metrics} \label{sec:distance}
We define the distances used to evaluate number, range, and Gaussian embeddings.  

\subsubsection{Numbers (absolute difference)}
For two scalars $x, y \in \mathbb{R}$, the distance is
\begin{equation}
d_{\text{num}}(x, y) = |x - y|
\end{equation}
\subsubsection{Ranges}
For two ranges $R_i = [a,b]$ and $R_j = [c,d]$, we consider two distances: 

Euclidean distance \cite{euclidean}:  Center and length are defined as $c_i=\tfrac{a+b}{2}, \;\ell_i=|b-a|$ and 
$c_j=\tfrac{c+d}{2}, \;\ell_j=|d-c|$, and the distance is
\begin{equation}
    d_{\text{CL}}(R_i, R_j) = \sqrt{(c_i - c_j)^2 + (\ell_i - \ell_j)^2}
\end{equation}

Jaccard-based Intersection over Union (IoU) distance \cite{niwattanakul2013using}:  
\begin{equation}
    d_{\text{IoU}}(R_i,R_j) = 1 - \frac{|R_i \cap R_j|}{|R_i \cup R_j|}
\end{equation}
where the intersection length is $\max(0, \min(b,d) - \max(a,c))$ and 
the union length is $(b-a) + (d-c) - |R_i \cap R_j|$. 

\subsubsection{Gaussians (2-Wasserstein distance)} 
For two Gaussians $\mathcal{N}(\mu_i,\sigma_i^2)$ and $\mathcal{N}(\mu_j,\sigma_j^2)$ the distance is \cite{peyre2019computational},  
\begin{equation}
    d_{\mathcal{W}_2}(i,j) = \sqrt{(\mu_i - \mu_j)^2 + (\sigma_i - \sigma_j)^2}
\end{equation}

\begin{figure}[htpb]
 \centering
  \includegraphics[width=0.6\columnwidth]{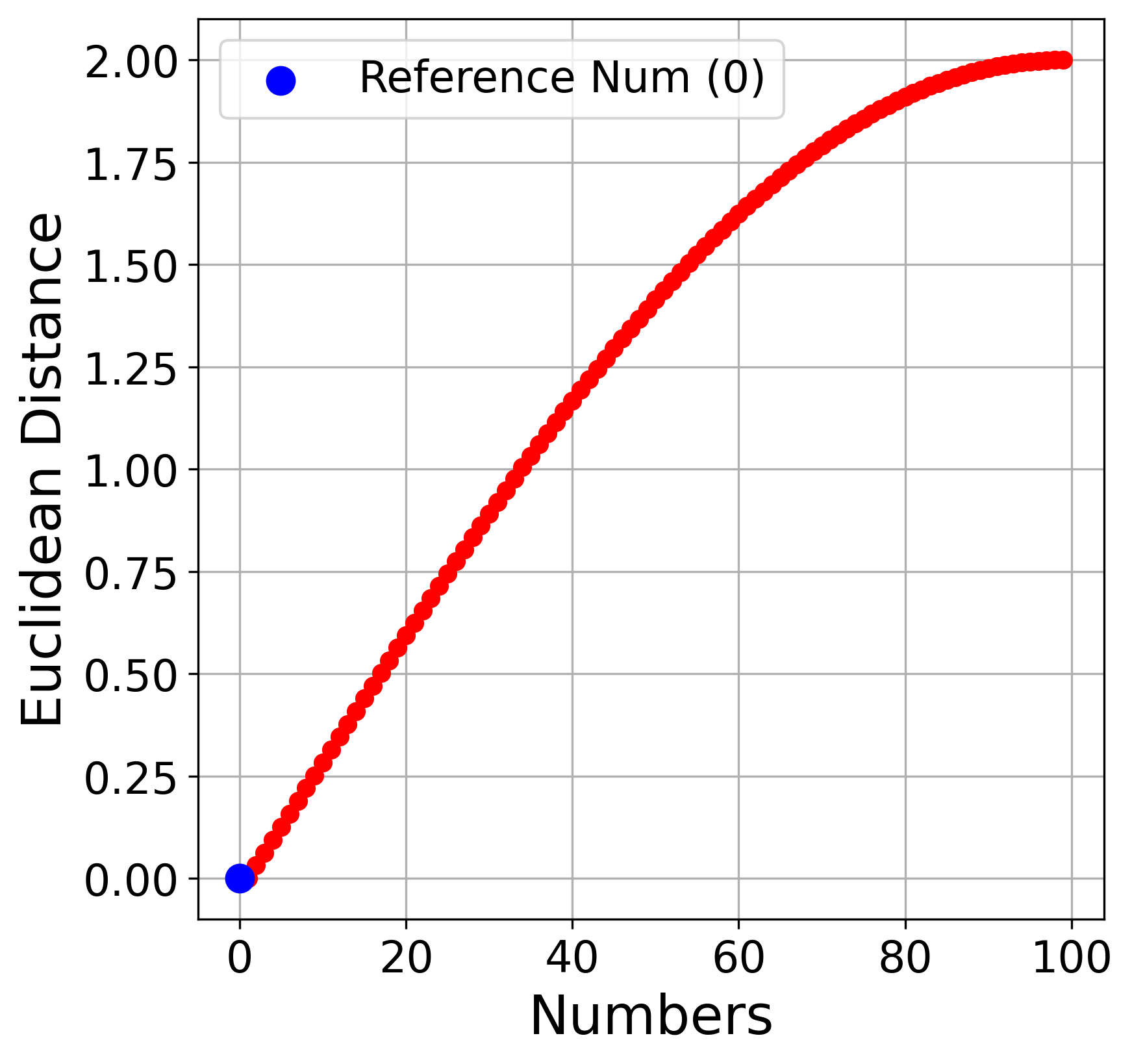}
 \caption{Euclidean Distance from Embedding of 0 to the Embeddings of Numbers from 0 to 100}.
     \vspace{-2ex}  
 \label{fig:dice_num}
\end{figure}

\subsection{\texttt{CONE} Embedding Components}
\subsubsection{Numbers}
In Figure \ref{fig:dice_num}, we plot the pairwise Euclidean distances for \texttt{CONE} embedding vectors corresponding to 0 and numbers from 0 to 100. As the numbers increase, the distances grow accordingly as expected, demonstrating \texttt{CONE}’s ability to capture numerical magnitude, effectively encoding both \textit{order} and \textit{interval} properties \cite{stevens1946theory}. The distance is monotonically increasing (larger numbers map farther away from 0), but starts demonstrating monotonic non-linearity with growth. Due to space, we show here the results for numbers from 0 to 100, however the same dynamics remains for much larger numbers (i.e. $\approx$1000) as well according to our experiments. We also evaluate whether embeddings preserve actual distances of numbers (scalars, ranges, and gaussians) by comparing these distances (Section~\ref{sec:distance}) with corresponding distances in the embedding space. We report \textit{Pearson’s} correlation coefficient ($r$) for linear association and \textit{Spearman’s} rank correlation coefficient ($\rho$) for rank-order consistency \citep{lee1988thirteen}. Results are summarized in Table~\ref{tab:dis_eval}. Quantitatively, correlating absolute differences of numbers (from 0–1000) with their corresponding embedding cosine distances yields weak correlation for standard encoder embeddings (e.g., BioBERT; $r\!\approx\!0.07$, $\rho\!\approx\!0.06$), while \texttt{CONE} achieves strong correlation ($r\!=\!0.989$, $\rho\!=\!0.798$). This shows that \texttt{CONE} preserves numerical magnitude, unlike standard encoders.

\subsubsection{Ranges}
\begin{figure}[t]
 \centering
  \includegraphics[width=0.9\columnwidth] {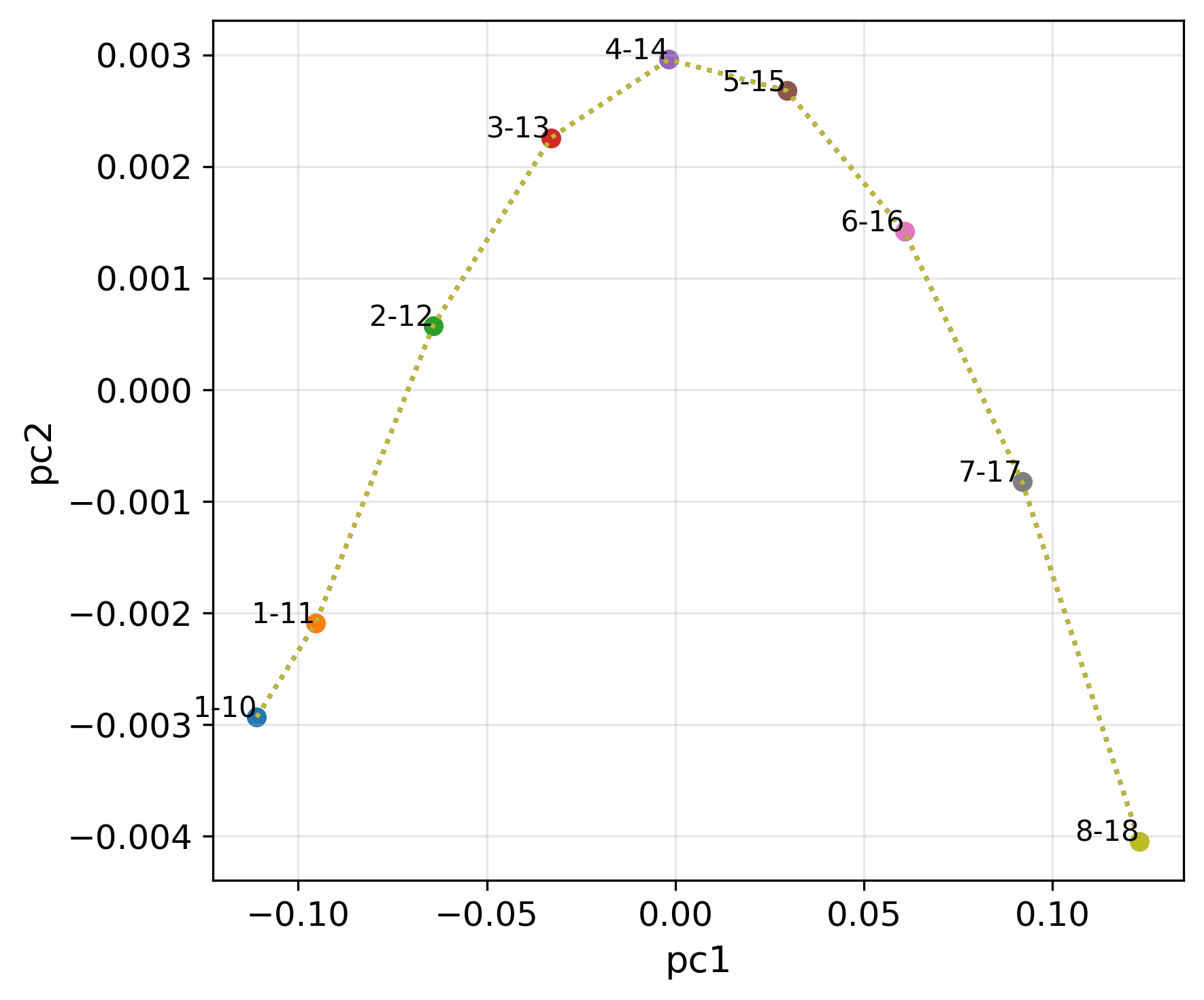}
 \caption{PCA-transformed \texttt{CONE} Embedding for Numerical Ranges $[1,20]$ without Units.}
 \vspace{-2ex}
 \label{fig:dice_range}
\end{figure}

We generated 1,000 random ranges, uniformly sampled from $[0,100]$, $[1,1000]$, and $[0,10000]$. Figure \ref{fig:dice_range} shows 2D scatter plot of PCA (Principal Component Analysis)-transformed \texttt{CONE} embeddings for randomly sampled ranges from $[1,20]$. We can see that \texttt{CONE} preserve the relative distance of ranges: similar ranges (e.g., ``\textit{1-10}” and ``\textit{1-11}”) are positioned close to each other, reflecting their proximity. Similarly, other adjacent ranges, such as ``\textit{2-12}” and ``\textit{3-13}” are also embedded nearby, which is the expected behavior. As the left boundary of the range increases, the distance between it and the non-overlapping ranges with the smaller lower boundary increases, which is also the expected correct behavior. For example, larger ranges, such as ``\textit{4-14}” and ``\textit{5-15}” are far from the smaller ranges such as ``\textit{1-10}” and ``\textit{1-11}”. Further, we evaluate two analytic distance notions (Section~\ref{sec:distance}): Euclidean ($d_{\text{CL}}$) \cite{euclidean} and Jaccard-based IoU distance ($d_{\text{IoU}}$) \cite{niwattanakul2013using}, correlating them with the embedding cosine distances. BioBERT exhibits moderate correlation on $d_{\text{CL}}$ ($r\!=\!0.398$, $\rho\!=\!0.355$) and $d_{\text{IoU}}$ ($r\!=\!0.267$, $\rho\!=\!0.0.248$), whereas \texttt{CONE} attains very high correlation on $d_{\text{CL}}$ ($r\!=\!0.997$, $\rho\!=\!0.786$) and substantially stronger alignment on $d_{\text{IoU}}$ ($r\!=\!0.498$, $\rho\!=\!0.465$; Table~\ref{tab:dis_eval}). These results indicate that the \texttt{CONE} embedding space encodes both range magnitudes and their relative distances accurately. 

\subsubsection{Gaussians}
We sampled $500$ gaussian distributions with $\mu \sim \mathcal{U}(0,100)$ and $\sigma \sim \mathcal{U}(1,10)$. We then compared embedding cosine distances against 2-Wasserstein distances \citep{peyre2019computational}. BioBERT embedding is inconsistent with actual expected distance ($r$ $\approx 0.04$). In contrast, \texttt{CONE} significantly improves correlation ($r\!=\!0.689$, $\rho\!=\!0.663$; Table~\ref{tab:dis_eval}), indicating that the embedding space captures these distributional properties, though performance remains below that of scalars and ranges.
\begin{figure}
    \centering
    \includegraphics[width=1\columnwidth]{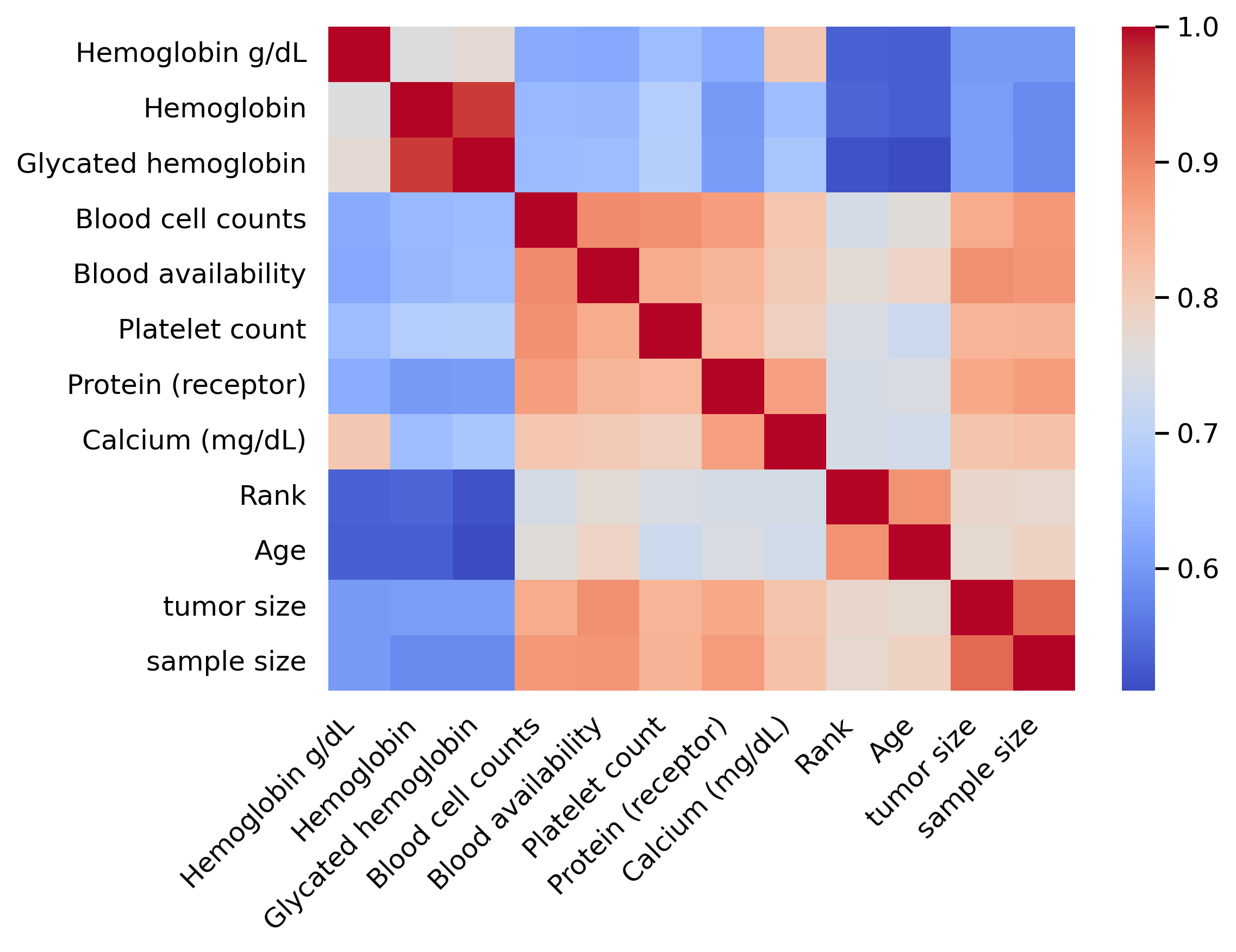}
    \caption{Attributes Similarity with \texttt{CONE} Embeddings.}
    \vspace{-2ex}  
    \label{fig:attribute}
\end{figure}
\subsubsection{Attributes}
To evaluate attribute-level semantic similarity, we sampled 100 attributes (i.e column names) from our dataset and compared their embedding distances using \texttt{CONE}.  Due to space constraints, we visualize 12 representative attributes in the heatmap shown in Figure \ref{fig:attribute}. The color intensity (blue to red) denotes pairwise similarity, where darker red indicates higher similarity (values close to 1.0) and blue indicates lower similarity. Similar attributes, such as \textit{Hemoglobin} and \textit{Glycated hemoglobin} have closer distance, as indicated by the red regions. This reflects their proximity in the embedding vector space, as expected from their sematic relatedness. On the other hand, different attributes, such as \textit{Hemoglobin} and \textit{Age} show lower similarity, represented by blue regions, indicating greater distance between their embeddings. This suggests that different attributes are appropriately far apart. Because this evaluation relies solely on attribute names, \textit{tumor size} and \textit{sample size} appear relatively close, despite referring to fundamentally different quantities (tumor measurements versus population counts). This behavior is characteristic of the pre-trained text encoders when numerical values and unit context are not considered. \texttt{CONE}’s composite embedding design addresses this limitation by integrating attribute, unit, and numeric components when full column data are available (see evaluation results in Table \ref{tab:baselien_comparison}). We also observe that units may appear either explicitly in attribute names (e.g., \textit{Hemoglobin g/dL}) or implicitly in column values. When only considering attribute similarity, \textit{Hemoglobin g/dL} and \textit{Hemoglobin} are somewhat farther apart, even though they represent similar attributes. This discrepancy is addressed by our composite embedding structure with 3 components, which extracts missing unit from column data and jointly encodes it with attribute semantics. Interestingly, some attributes, such as \textit{Hemoglobin g/dL} and \textit{Calcium (mg/dL)} are relatively close in similarity due to their common unit (\textit{g/dL} or \textit{mg/dL}), even though they represent different biological measures. This is also addressed by having the 3 components together in a single vector. Concatenating the unit embeddings with attribute embeddings will further enhance our model’s ability to understand and distinguish numerical values correctly and precisely. We compared attribute similarity results from \texttt{CONE} with those from a fine-tuned BioBERT and observed no notable difference, confirming that \texttt{CONE} retains its ability to encode textual data effectively while also capturing numerical and unit-level semantics.


\begin{figure}
    \centering
    \includegraphics[width=1\linewidth]{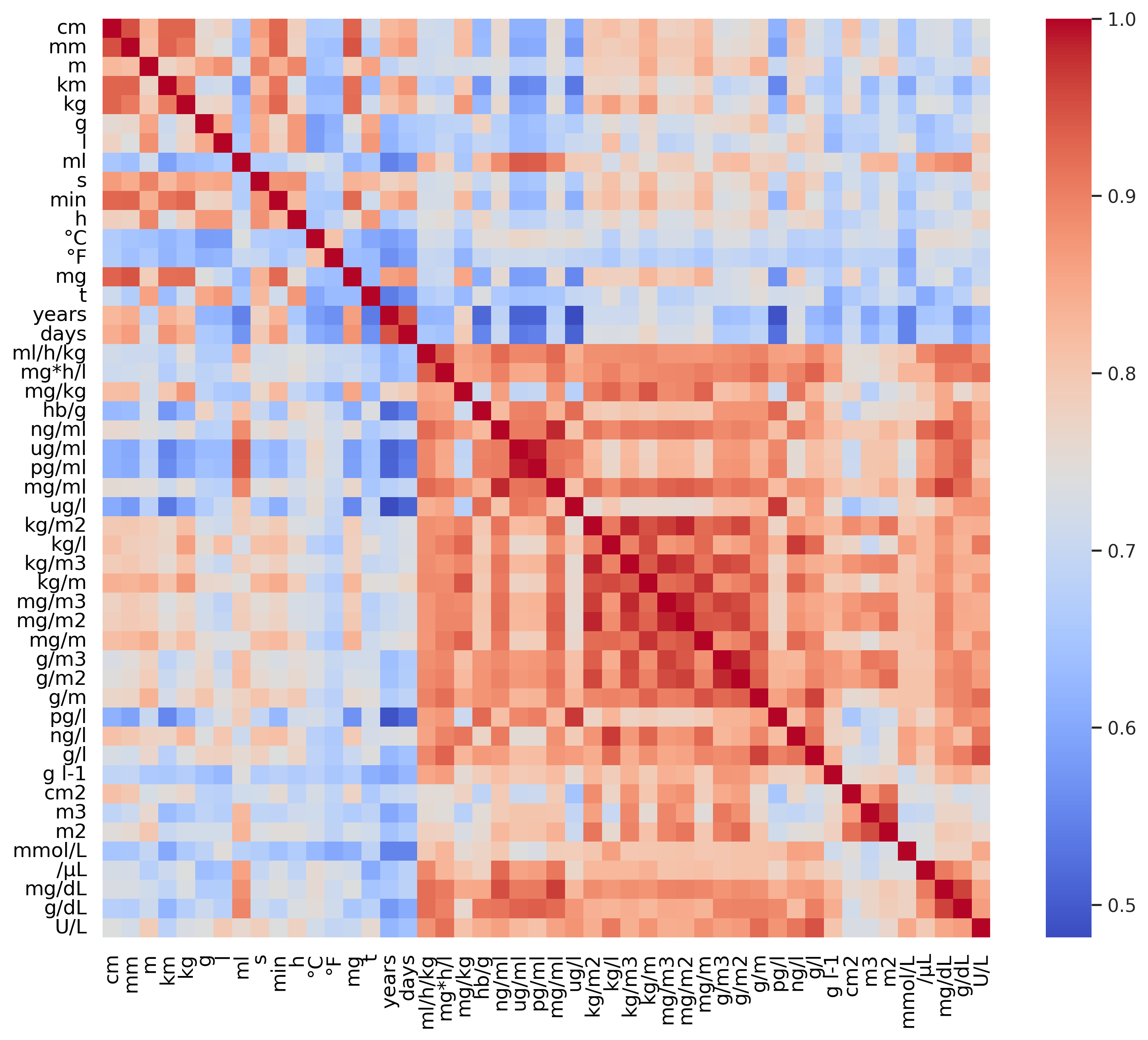}
    \caption{Units Similarity with \texttt{CONE} Embeddings.}
    \vspace{-2ex}  
    \label{fig:units}
\end{figure}

\subsubsection{Units}
To evaluate our unit embeddings, we sampled 50 diverse units from our dataset and visualized pairwise cosine distances. Figure \ref{fig:units} shows the distance between different units. Units of length, such as \textit{cm}, \textit{mm}, \textit{km}, and \textit{m} are highly similar to each other (indicated by red regions), reflecting their relatedness in terms of physical measurement. Similarly, weight-related units such as \textit{kg} and \textit{g} are closely located, as expected, since they measure the same physical quantity. Time-related units, such as \textit{h} (hours), \textit{min} (minutes), and \textit{s} (seconds), also show high similarity, indicating their connection through temporal measurements. Units like \textit{°C} (degrees Celsius) and \textit{°F} (degrees Fahrenheit) exhibit moderate similarity, reflecting their shared context as temperature units but with different scales. Units commonly used in biological and chemical measurements, such as \textit{mg/mL}, \textit{ng/mL}, and \textit{pg/mL} show significant similarity, which is expected because they represent quantities related to mass concentration, just on different scales. On the other hand, units like \textit{mg/kg} and \textit{mg/dL} are relatively distinct from units, such as \textit{mm} and \textit{cm} as indicated by the blue regions. This distinction is logical, as these units measure completely different physical quantities (\textit{mass}, \textit{length}, and \textit{concentration}). This demonstrates the role of the unit embedding component.

\begin{figure}[t]
 \centering
  \includegraphics[width=1.0\columnwidth]
   {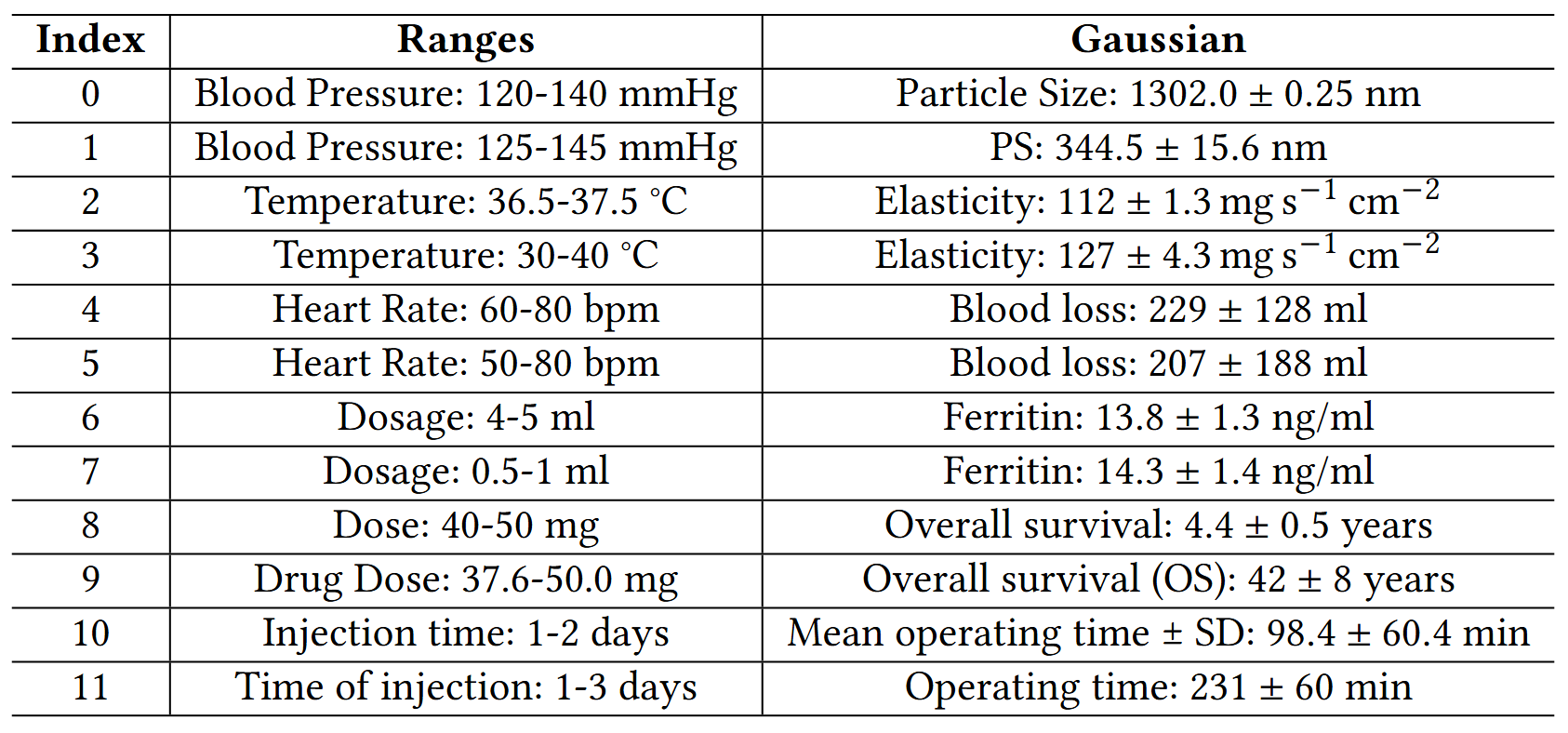}
 \caption{Sample Numerical Ranges and Gaussians.}
     \vspace{-2ex}  
 \label{fig:sample-composite-range-gaussian}
\end{figure}

\begin{figure}[htpb]
    \centering
    \begin{subfigure}[b]{0.48\linewidth}
        \centering
        \includegraphics[width=\linewidth]{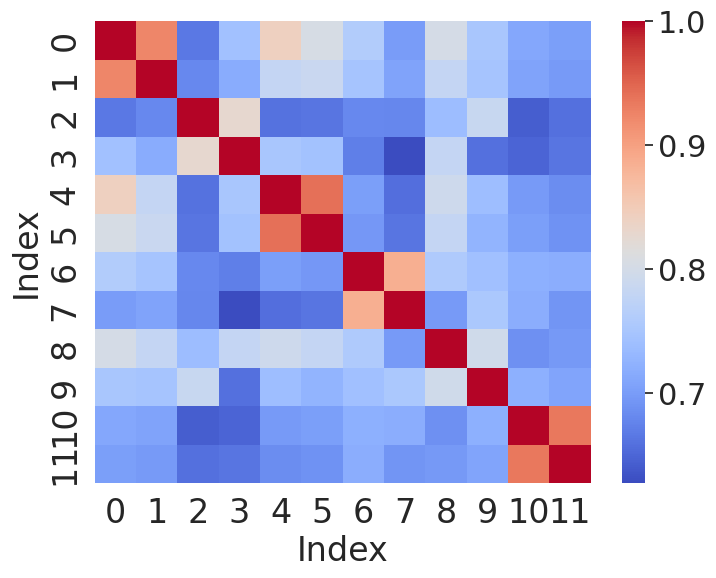}
        \caption{}  
        \label{fig:range}
    \end{subfigure}
    \hfill
    \begin{subfigure}[b]{0.48\linewidth}
        \centering
        \includegraphics[width=\linewidth]{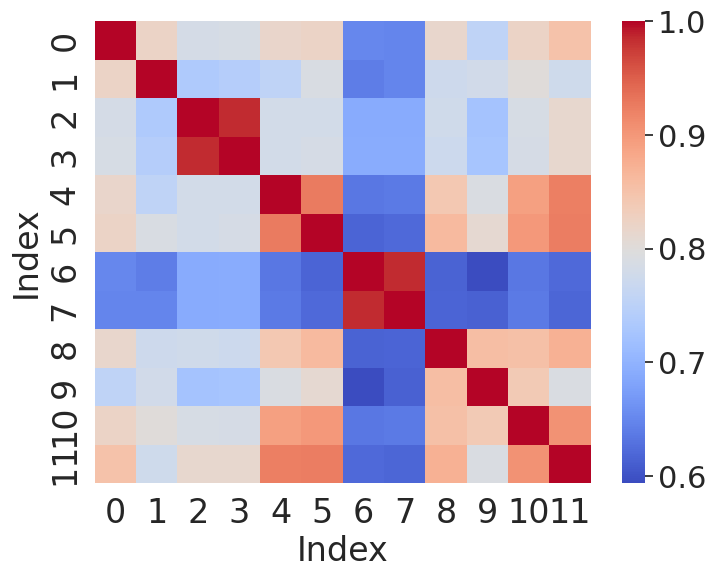}
        \caption{}  
        \label{fig:gaussian}
    \end{subfigure}
    \captionsetup{skip=2pt} 
    \caption{Distance Between \texttt{CONE} Composite Embedding for Numerical (a) Ranges and (b) Gaussians with Attributes and Units in Figure \ref{fig:sample-composite-range-gaussian}.}
    \vspace{-2ex}  
    \label{fig:heatmap_range_gaussian}
        
\end{figure}
Next, to validate our composite embedding structure for ranges, we show that it captures composite semantics by incorporating structural context (attribute, unit, range center and length). Figure \ref{fig:sample-composite-range-gaussian} presents sampled ranges, and gaussians along with their associated attributes and units from our dataset. In Figure \ref{fig:heatmap_range_gaussian}(a) heatmap ranges with similar structural context (e.g., index 0, \textit{Blood Pressure: 120-140 mmHg} and index 1, \textit{Blood Pressure: 125-145 mmHg}) have high similarity, as they are closer to each other in the composite embedding space. 
Conversely, while indices 6 (\textit{Dosage: 4-5 ml}), 7 (\textit{Dosage: 0.5-1 ml}), and 8 (\textit{Dose: 40-50 mg}) share a similar attribute, the different unit of index 8 places it farther from indices 6 and 7 in the composite embedding space, which is correct. This demonstrates that the composite embedding mitigates complete domination by textual similarity alone; otherwise, they would have been much closer. The embeddings of indices 6 and 7 are closer (0.89) than those of indices 8 and 9 (0.8), mainly due to minor attribute wording differences (\textit{Dose} vs. \textit{Drug dose}). Nevertheless, the similarity score between indices 8 and 9 is higher than the score when either index 8 or index 9 is compared with any other index in Figure \ref{fig:sample-composite-range-gaussian}. Importantly, when attributes are removed, numerical ranges preserve proportional distance (Table \ref{tab:dis_eval}).

Similarly, Figure \ref{fig:heatmap_range_gaussian}(b) heatmap shows the distance between composite embeddings of gaussians in Figure \ref{fig:sample-composite-range-gaussian}. Gaussian pairs with similar contextual structures (e.g., index 6, \textit{Ferritin: $13.8\pm1.3$ ng/ml} and index 7, \textit{Ferritin: $14.3\pm1.4$ ng/ml}) exhibit a closer distance compared to other pairs. We also observe potential groupings of closely related clinical parameters, such as \textit{Blood loss} (indices: 4, 5), \textit{Overall survival} (indices:8, 9) and \textit{operating time}. However, the distances between these groups are smaller compared to those between the corresponding similar index pairs.

\begin{figure}[t]
  \centering
  \begin{subfigure}[b]{0.47\textwidth}
    \centering
    \includegraphics[width=\textwidth]{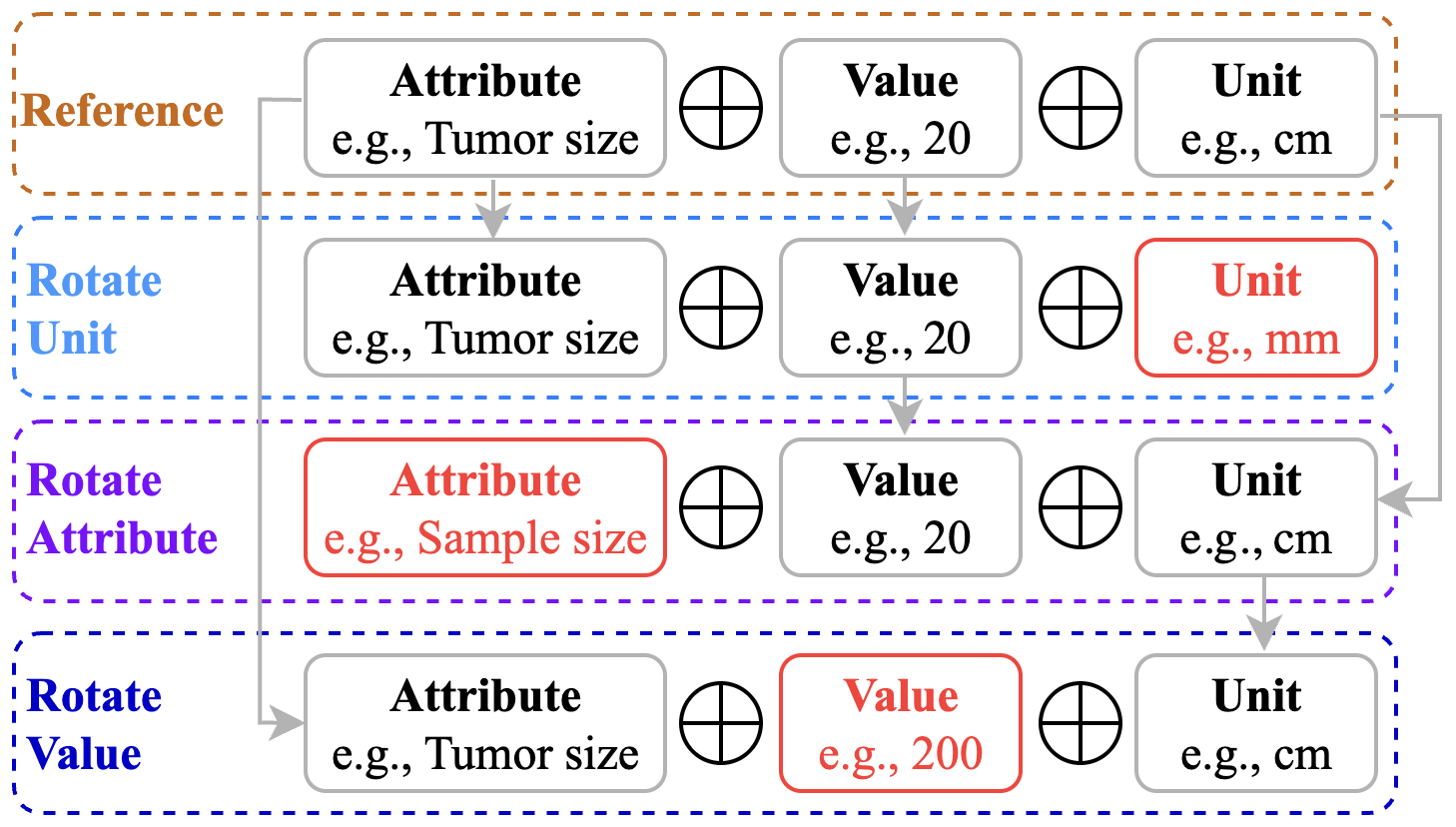}
    \caption{}
    \label{fig:composite_rotate_component_a}
  \end{subfigure}
  \hfill
  \begin{subfigure}[b]{0.47\textwidth}
    \centering
    \includegraphics[width=\textwidth]{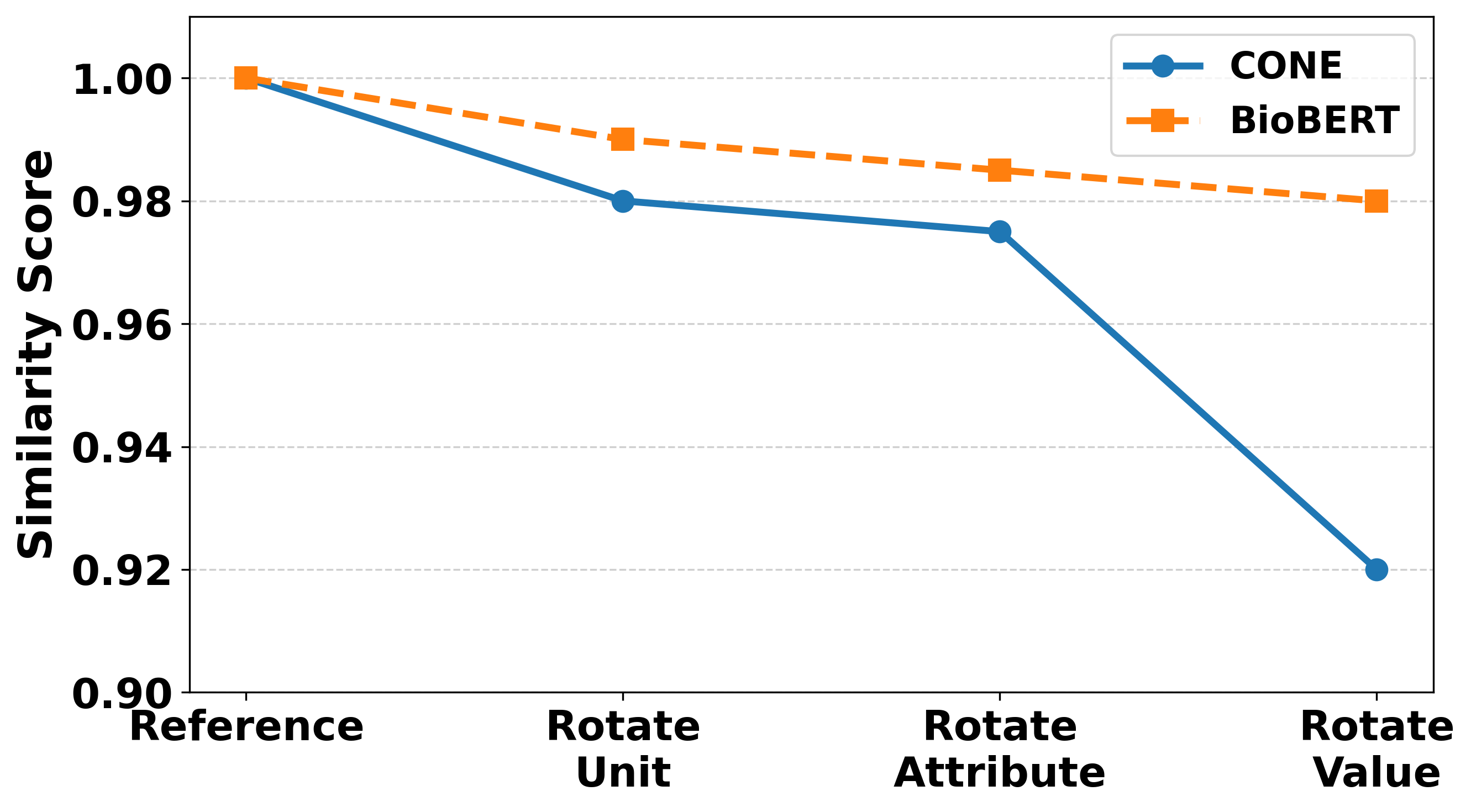}
    \caption{}
    \label{fig:composite_rotate_component_b}
  \end{subfigure}
\captionsetup{skip=2pt} 
  \caption{ Effect of "Rotating" (see Section \ref{sec:rotate}) Individual Components in \texttt{CONE} Composite Embeddings: (a) Illustration of Component-wise "Rotation"; (b) Change in Cosine Similarity Relative to the Original Composite Embedding.}
  \vspace{-2ex}  
  \label{fig:composite_rotate_component}
\end{figure}

\subsubsection{Ablation study of Composite Components} \label{sec:rotate}
Finally, we analyze the significance of each component in our composite embeddings by observing how the similarity score (embedding distance) changes when one component is perturbed while other staying the same. We refer to this process as "rotation". Specifically: (1) keeping the attribute name and value fixed, we "rotate" the unit; (2) keeping the unit and value fixed, we "rotate" the attribute; and (3) keeping the attribute and unit fixed, we "rotate" the value. Consider a reference example of a tuple (\textit{attribute}, \textit{value}, and \textit{unit}) as (`\textit{Tumor size}', `\textit{20}', `\textit{cm}') (Figure \ref{fig:composite_rotate_component}(a) illustrates an example of a composite input reference, while Figure \ref{fig:composite_rotate_component}(b) shows the change in cosine similarity distance relative to the original embedding when each component is "rotated"). Rotating the unit (\textit{cm}~$\rightarrow$~\textit{mm}) results in a slight similarity drop from 1.0 to 0.98. Rotating the attribute (\textit{Tumor size}~$\rightarrow$~\textit{Sample size}) leads to a drop to 0.975, while rotating the value (\textit{20}~$\rightarrow$~\textit{200}) causes the largest decrease, down to 0.92. This analysis reveals that the value component has the most significant impact on the embedding, indicating its critical role in preserving semantic meaning. The attribute and unit have a moderate influence. These component-wise variations provide a holistic view of the embedding's sensitivity and further validate our design: each component (\textit{attribute}, \textit{value}, and \textit{unit}) in the composite embedding structure uniquely contribute to capturing semantics of complex structured numerical data. It is important to note that the relative contribution of each component may vary depending on the specific structural context. In Figure \ref{fig:composite_rotate_component}(b) we can see that \texttt{CONE} captures semantic structure more effectively than BioBERT, primarily due to its integration of numerical context in the embedding. 

\section{Experimental Evaluations} \label{sec:evaluation}
To examine the numerical properties of \texttt{CONE}, we consider three synthetic tasks of \textit{list maximum}, \textit{decoding}, and \textit{addition}, as proposed by \cite{wallace2019nlp}. We then evaluate performance on the DROP QA benchmark \cite{dua2019drop}, which requires numerical reasoning skills such as counting, sorting, and addition. Finally, we assess the quality of the composite embedding representation by measuring its performance on two popular downstream tasks over structured data, using large-scale datasets, and compare it against SOTA baselines.

\begin{table*}[t]
\centering
\caption{Experimental Results on List Maximum, Decoding, and Addition.}
\label{tab:three_numercy_task}
\resizebox{\textwidth}{!}{%
\begin{tabular}{l|ccccc|ccccc|ccccc}
\hline
 & \multicolumn{5}{c|}{\textbf{List Maximum (accuracy)}} & \multicolumn{5}{c|}{\textbf{Decoding (RMSE)}} & \multicolumn{5}{c}{\textbf{Addition (RMSE)}} \\
\textit{Integer range} & $[0, 10^2]$ & $[0, 10^3]$ & $[0, 10^4]$ & $[0, 10^5]$ & $[0, 10^6]$ & $[0, 10^2]$ & $[0, 10^3]$ & $[0, 10^4]$ & $[0, 10^5]$ & $[0, 10^6]$ & $[0, 10^2]$ & $[0, 10^3]$ & $[0, 10^4]$ & $[0, 10^5]$ & $[0, 10^6]$ \\ \hline
Random vectors & 0.17\textsubscript{$\pm$0.030} & 0.22\textsubscript{$\pm$0.030} & 0.20\textsubscript{$\pm$0.035} & 0.18\textsubscript{$\pm$0.038} & 0.16\textsubscript{$\pm$0.039} & 30.1\textsubscript{$\pm$1.80} & 295.88\textsubscript{$\pm$15} & 2888.62\textsubscript{$\pm$42} & 28k\textsubscript{$\pm$1k} & 280k\textsubscript{$\pm$5k} & 41.80\textsubscript{$\pm$0.50} & 412.33\textsubscript{$\pm$12} & 4400.39\textsubscript{$\pm$55} & 40k\textsubscript{$\pm$2k} & 405k\textsubscript{$\pm$8k} \\
BioBERT & 0.94\textsubscript{$\pm$0.025} & 0.63\textsubscript{$\pm$0.028} & 0.50\textsubscript{$\pm$0.022} & 0.47\textsubscript{$\pm$0.030} & 0.42\textsubscript{$\pm$0.032} & 3.28\textsubscript{$\pm$0.10} & 29.7\textsubscript{$\pm$0.80} & 434\textsubscript{$\pm$10} & 4.1k\textsubscript{$\pm$0.2k} & 40.5k\textsubscript{$\pm$2k} & 4.61\textsubscript{$\pm$0.20} & 69.2\textsubscript{$\pm$2.10} & 461\textsubscript{$\pm$12} & 4.2k\textsubscript{$\pm$0.3k} & 42.1k\textsubscript{$\pm$3k} \\ 
DICE & 0.97\textsubscript{$\pm$0.002} & \textbf{0.96\textsubscript{$\pm$0.001}} & 0.96\textsubscript{$\pm$0.002} & 0.95\textsubscript{$\pm$0.002} & 0.94\textsubscript{$\pm$0.003} & 0.45\textsubscript{$\pm$0.03} & 0.83\textsubscript{$\pm$0.03} & 3.02\textsubscript{$\pm$0.08} & 5.42\textsubscript{$\pm$0.15} & 11.2\textsubscript{$\pm$0.4} & \textbf{0.64\textsubscript{$\pm$0.02}} & 3.80\textsubscript{$\pm$0.15} & 29.2\textsubscript{$\pm$1.2} & 55.1\textsubscript{$\pm$2.1} & 108.4\textsubscript{$\pm$4} \\
NTL-WAS & 0.95\textsubscript{$\pm$0.003} & 0.94\textsubscript{$\pm$0.003} & 0.92\textsubscript{$\pm$0.002} & 0.91\textsubscript{$\pm$0.002} & 0.88\textsubscript{$\pm$0.003} & \textbf{0.30\textsubscript{$\pm$0.03}} & \textbf{0.49\textsubscript{$\pm$0.12}} &2.75\textsubscript{$\pm$0.15} & 18.4\textsubscript{$\pm$1.20} & 45.0\textsubscript{$\pm$2.5} & 1.26\textsubscript{$\pm$0.03} & 4.25\textsubscript{$\pm$0.25} & 36.4\textsubscript{$\pm$1.8} & 82.4\textsubscript{$\pm$4.1} & 158.2\textsubscript{$\pm$8} \\
\hline
\texttt{CONE} & \textbf{0.98\textsubscript{$\pm$0.002}} & 0.95\textsubscript{$\pm$0.002} & \textbf{0.97\textsubscript{$\pm$0.001}} & \textbf{0.96\textsubscript{$\pm$0.001}} & \textbf{0.95\textsubscript{$\pm$0.003}} & 0.42\textsubscript{$\pm$0.02} & 0.73\textsubscript{$\pm$0.03} & \textbf{2.71\textsubscript{$\pm$0.05}} & \textbf{4.89\textsubscript{$\pm$0.10}} & \textbf{9.85\textsubscript{$\pm$0.3}} & \textbf{0.64\textsubscript{$\pm$0.02}} & \textbf{2.18\textsubscript{$\pm$0.08}} & \textbf{26.10\textsubscript{$\pm$0.9}} & \textbf{48.9\textsubscript{$\pm$1.8}} & \textbf{95.1\textsubscript{$\pm$3}} \\ \hline
\end{tabular}%
}
\end{table*}

\begin{table}[t]
\caption{Results (in \%) on the DROP Dataset\textsuperscript{\ref{fn:commonfootnote}}}.
\vspace{-2ex}
\label{tab:drop}
\centering
\footnotesize
\begin{tabular}{lcccc}
\hline
\textbf{Method} & \multicolumn{2}{c}{\textbf{Dev}} & \multicolumn{2}{c}{\textbf{Test}} \\
\cline{2-3}\cline{4-5}
 & \textbf{EM} & \textbf{F1} & \textbf{EM} & \textbf{F1} \\
\hline
NumNet \cite{ran2019numnet} & 83.34$\pm$0.2 & 86.00$\pm$0.38 & 83.40$\pm$0.18 & 86.42$\pm$0.4 \\

NC-BERT \cite{kim2022exploiting} &  75.18$\pm$1.2& 77.68$\pm$0.75 & 76.02$\pm$0.9 & \textcolor{red}{77.91$\pm$0.8} \\
AeNER \cite{yarullin2023numerical} & \textbf{83.72} & 86.56  & 83.69 & 86.98 \\
\hline
\texttt{CONE} & 83.71$\pm$0.12 & \textbf{86.70$\pm$0.2} & \textbf{83.74$\pm$0.12} & \textbf{\textcolor{red}{87.28$\pm$0.3}} \\
\hline
\end{tabular}
\end{table}

\subsection{Experimental Setup}

\noindent\textbf{Implementation:} We experimented on a server with a 112-core Intel Xeon Platinum 8180 2.50 GHz CPU, 3 NVIDIA V100 GPUs, 1.5 TB of RAM, and 16 TB of disk space. For the encoder, we used BioBERT-baseuncased with a hidden layer = 12, hidden units = 768, and attention heads = 12. We used the Adam optimizer with a learning rate of \( 2 \times 10^{-5} \), selected from \( \{ 1.5 \times 10^{-5}, 2 \times 10^{-5} \} \), and a batch size of 12. The maximum sequence length was set to 512. We took the vocabulary and pre-trained embedding weights from BioBERT and fine-tuned it on our training datasets. The configuration of BioBERT is aligned with $BERT_{\text{BASE}}$ model. We trained for 50,000 steps ($\approx$ 3 days), with hyperparameters specified above; the masked numeral prediction loss stabilized well before the final step. The training set was comprised of table tuples/columns. Each tuple/column was tokenized, embedded, and encoded following \cite{lee2020biobert}. To address token-length constraints, inputs exceeding 512 tokens were chunked, with [CLS] placed at the start of each tuple/column and [SEP] tokens inserted between cells.

\subsection{Numerical Reasoning}
\subsubsection{Probing Numeracy of Embeddings}
We first examine the numerical properties of \texttt{CONE} by evaluating it on three numeracy tasks - list maximum, decoding, and addition \cite{wallace2019nlp}. Each task uses the final contextualized numeral embedding $E^O$ from \texttt{CONE} and results are averaged over three runs with different random seeds. In list maximum, the model identifies the index of the largest value among five input numbers using a two-layer multilayer perceptron (MLP). Decoding involves regressing a single number’s embedding back to its original value using a five-layer MLP. For addition, the model predicts the sum of two numbers given their embeddings using a similar five-layer network. We sample integers from ranges $[0, 10^2]$ to $[0, 10^6]$ and use the 80\%/20\% train-test split, extending the evaluation protocol in \cite{wallace2019nlp} to include a larger range of values. Our experiments on even wider ranges (e.g., $10^8$) yielded similar performance trends, indicating that the relative advantage of \texttt{CONE} remains stable at scale. Hence, the range $[0-10^6]$ is adequate as it also covers the majority of numerical magnitudes encountered in common numeracy benchmarks. The list maximum task is evaluated using accuracy, while both decoding and addition tasks are evaluated using root mean squared error (RMSE). The results in Table \ref{tab:three_numercy_task} (mean and standard deviation) demonstrate \texttt{CONE} outperforms other baselines in most cases, indicating its highly accurate comprehension of numeracy with the composite embeddings. To evaluate the statistical significance, we used paired bootstrap resampling over the test set ($n=1000$) \cite{accuracy1986bootstrap}, constructing 95\% confidence intervals (CI) for the performance difference between strong baselines (DICE and NTL-WAS) and \texttt{CONE}. Due to space constraints, we omit the full set of CI results and instead report a representative example: on the addition task at $[0,10^6]$, \texttt{CONE} achieves an RMSE improvement of $\Delta=13.3$ over DICE with 95\% CI of $[9.8,16.7]$, which excludes zero and thus indicates a statistically significant gain. Different methods excel on different tasks based on how they encode numerical properties. For list maximum task, methods with explicit magnitude encoding (e.g. our method) achieve the strongest performance when direct comparison of numeric magnitudes is required. For the decoding task, NTL-WAS \cite{zausinger2025regress} performs well due to its digit-level reconstruction loss; our method matches or approaches its performance across ranges. For addition task, our method's explicit positional encoding and magnitude reconstruction loss (i.e. Equation \ref{loss}) contribute to improved performance.

\footnotetext[1]{The best results are in bold, and the largest delta $\Delta$ is highlighted in red. \label{fn:commonfootnote}}

\subsubsection{Numerical Reasoning on DROP}
In this section, we evaluate the proposed \texttt{CONE} model on the DROP dataset by measuring accuracy on questions that require numerical reasoning. Following \cite{ran2019numnet, kim2022exploiting, yarullin2023numerical}, we report Exact Match (EM) and F1 score, and compare against encoder-only baselines with numerical reasoning capabilities -- NumNet \cite{ran2019numnet}, NC-BERT \cite{kim2022exploiting} and AeNER \cite{yarullin2023numerical} -- as discussed in Section \ref{sec:related_work}. All experiments are averaged over three random seeds, and we report the mean and standard deviation. Since AeNER does not provide publicly available code, we cite its results directly from the original paper.
Table~\ref{tab:drop} shows that \texttt{CONE} improves over NumNet by $0.34$ EM / $0.86$ F1, and over NC-BERT by $7.72$ EM / $9.37$ F1 on the test set. Compared to AeNER, \texttt{CONE} slightly surpasses it with 0.05 EM / 0.3 F1. Across three random seeds, \texttt{CONE} demonstrates a statistically significant improvement over NumNet on Test F1 (paired $t$-test, $p < 0.01$), confirming the effectiveness of our numerical embedding design.

\subsection{Column and Tuple Matching} \label{subsec:cmtm}
This downstream task evaluates whether our composite embeddings can accurately identify semantically and numerically similar columns and tuples. Given a reference column or tuple, the objective is to retrieve its correct matches based on the embedding similarity.

\subsubsection{Baseline Methods}

We compare against several SOTA methods, TAPAS \cite{herzig2020tapas}, NumNet \cite{ran2019numnet}, and NC-BERT \cite{kim2022exploiting}, each fine-tuned on our datasets. 
TAPAS is a transformer model for table reasoning. NumNet \cite{ran2019numnet} constructs a number comparison graph that encodes the relative magnitude information between numbers on directed edges. We used NumNet+ model in our experiment, an enhanced version of NumNet, which uses a pre-trained RoBERTa model \cite{liu2019roberta}. NC-BERT \cite{kim2022exploiting} incorporates DICE as a loss function to revive the magnitude characteristics of numbers in the representations. 
NC-BERT is encoder-decoder model. Therefore, for unbiased comparison we use only encoder output for our experiment.

We also evaluated four general-purpose retrieval embedding models: BGE-M3 \cite{chen2024m3}, Stella (Stella-EN-1.5B-v5) \cite{zhang2024jasper}, Qwen3 (Qwen3-Embedding-4B) \cite{zhang2025qwen3}, and KaLM (KaLM-Embedding-Gemma3-12B-2511) \cite{hu2025kalmembedding, zhao2025kalmembeddingv2}. These models are widely used for semantic retrieval and similarity matching across heterogeneous text corpora. In addition, we compared \texttt{CONE} against several SOTA schema matching (SM) approaches, including the traditional method SimFlooding \cite{melnik2002similarity}, as implemented in Valentine \cite{koutras2021valentine}; ISResMat \cite{du2024situ}; the recent approach Magneto \cite{liu2025magneto}; and the tabular foundation model CARTE \cite{kim2024carte}. CARTE transforms tabular rows into graph representations to generate context-aware embeddings using a pretrained graph-attentional transformer. SimFlooding is a graph-based method that computes schema matches via fixed-point computation, while ISResMat uses contrastive learning and fine-tuned BERT.  However, both SimFlooding and ISResMat consistently performed significantly worse than other approaches across all column/tuple matching
experiments, so we exclude them from the reported results for clarity. Magneto performs candidate retrieval using a small language model (SLM) and reranking using an LLM. We evaluated Magneto in both zero-shot and fine-tuned configurations. Since the fine-tuned variant consistently outperformed the zero-shot setting, we report only the fine-tuned results for brevity.

\begin{table*}[t]
\caption{Average Recall@10 (R), MAP@10 and MRR@10 for Column Matching (CM) and Tuple Matching (TM)\textsuperscript{\ref{fn:commonfootnote}}.}
\label{tab:baselien_comparison}
\centering
\footnotesize
\begin{tabular}{l|l|ccc|ccc|ccc|ccc|ccc}
\hline
\textbf{Method} & \textbf{Task} & \multicolumn{3}{c|}{\textbf{CancerKG}} & \multicolumn{3}{c|}{\textbf{CovidKG}} & \multicolumn{3}{c|}{\textbf{WebTable}} & \multicolumn{3}{c|}{\textbf{CIUS}} & \multicolumn{3}{c}{\textbf{SAUS}} \\
& & R & MAP & MRR & R & MAP & MRR & R & MAP & MRR & R & MAP & MRR & R & MAP & MRR \\
\hline

\multirow{2}{*}{TAPAS} & CM & 0.783 & 0.728 & 0.742 & 0.734 & 0.669 & 0.688 & 0.800 & 0.736 & 0.753 & 0.820 & 0.771 & 0.789 & 0.820 & 0.774 & 0.787 \\
                       & TM & 0.736 & 0.683 & 0.701 & 0.700 & 0.642 & 0.658 & 0.800 & 0.733 & 0.745 & 0.800 & 0.751 & 0.764 & 0.800 & 0.752 & 0.767 \\
\hline
\multirow{2}{*}{NumNet} & CM & 0.650 & \textcolor{red}{0.581} & \textcolor{red}{0.596} & 0.700 & 0.628 & 0.641 & \textcolor{red}{0.700} & 0.612 & 0.628 & 0.758 & 0.686 & 0.703 & 0.780 & 0.707 & 0.724 \\
                        & TM & 0.700 & \textcolor{red}{0.619} & 0.634 & 0.667 & 0.598 & 0.611 & \textcolor{red}{0.723} & \textcolor{red}{0.629} & \textcolor{red}{0.639} & 0.750& 0.675 & 0.686 & 0.700 & 0.624 & 0.641 \\
\hline
\multirow{2}{*}{NC-BERT} & CM & 0.792 & 0.758 & 0.768 & 0.750 & 0.694 & 0.712 & 0.850 & 0.788 & 0.795 & 0.850 & 0.792 & 0.809 & 0.850 & 0.793 & 0.804 \\
                         & TM & 0.767 & 0.723 & 0.737 & 0.735 & 0.686 & 0.697 & 0.800 & 0.739 & 0.748 & 0.850 & 0.794 & 0.804 & 0.800 & 0.743 & 0.754 \\
\hline
\multirow{2}{*}{Magneto}     & CM & 0.784 & 0.741 & 0.770 & 0.724 & 0.685 & 0.718 & 0.838 & 0.796 & 0.802 & 0.820 & 0.760 & 0.800 & 0.810 & 0.762 & 0.796 \\
                         & TM & 0.742 & 0.691 & 0.738 & 0.700 & 0.662 & 0.675 & 0.780 & 0.706 & 0.717 & 0.800 & 0.751  & 0.764 & 0.750 & 0.712 & 0.725 \\
\hline
\multirow{2}{*}{CARTE }     & CM & 0.750 & 0.690 & 0.698 & 0.730 & 0.692 & 0.682 & 0.850 & 0.802 & 0.800 & 0.825 & 0.780 & 0.794 & 0.817 & 0.769 & 0.782 \\
                         & TM & 0.754 & 0.706 & 0.722 & 0.700 & 0.633 & 0.676 & 0.825 & 0.764 & 0.777 & 0.820 & 0.768 & 0.780 & 0.800 & 0.743 & 0.754 \\
\hline
\multirow{2}{*}{BGE-M3}     & CM & 0.795 & 0.761 & 0.771 & 0.750 & 0.713 & 0.726 & 0.902 & 0.806 & 0.811 & 0.853 & 0.795 & 0.802 & 0.852 & 0.795 & 0.802 \\
                         & TM & 0.796 & 0.758 & 0.769 & 0.740 & 0.703 & 0.716& 0.850 & 0.797 & 0.807 & 0.850 & 0.794 & 0.804 & 0.800 & 0.755 & 0.766 \\
\hline
\multirow{2}{*}{Stella}     & CM & 0.798 & 0.760 & 0.773 & 0.750 & 0.712 & 0.726 & 0.900 & 0.806 & 0.813 & 0.855 & 0.798 & 0.810 & 0.852 & 0.795 & 0.806 \\
                         & TM & 0.798 & 0.760 & 0.771 & 0.742 & 0.705 & 0.718 & 0.850 & 0.797 & 0.807 & 0.850 & 0.794 & 0.804 & 0.810 & 0.759 & 0.766 \\
\hline
\multirow{2}{*}{Qwen3}     & CM & 0.800 & 0.770 & 0.781 & 0.754 & 0.716 & 0.729 & 0.910 & 0.810 & 0.820 & 0.855 & 0.798 & 0.810 & 0.852 & 0.797 & 0.806 \\
                         & TM & 0.800 & 0.762 & 0.774 & 0.745 & 0.708 & 0.721 & 0.852 & 0.799 & 0.810 & 0.857 & 0.803 & 0.813 & 0.815 & 0.761 & 0.772 \\
\hline
\multirow{2}{*}{KaLM}     & CM & 0.800 & 0.770 & 0.781 & 0.760 & 0.722 & 0.735 & 0.910 & 0.810 & 0.820 & 0.860 & 0.804 & 0.818 & 0.856 & 0.801 & 0.811 \\
                         & TM & 0.810 & 0.775 & 0.786 & 0.750 & 0.713 & 0.726 & 0.858 & 0.806 & 0.816 & 0.860 & 0.806 & 0.816 & 0.815 & 0.765 & 0.772 \\
\hline
\multirow{2}{*}{\textbf{\texttt{CONE}}} & CM & \textbf{0.833} & \textbf{\textcolor{red}{0.817}} & \textbf{\textcolor{red}{0.831}} & \textbf{0.800} & \textbf{0.782} & \textbf{0.798} & \textbf{\textcolor{red}{0.950}} & \textbf{0.842} & \textbf{0.858} & \textbf{0.900} & \textbf{0.851} & \textbf{0.864} & \textbf{0.900} & \textbf{0.849} & \textbf{0.861} \\
                                        & TM & \textbf{0.867} & \textbf{\textcolor{red}{0.844}} & \textbf{0.853} & \textbf{0.800} & \textbf{0.781} & \textbf{0.792} & \textbf{\textcolor{red}{0.900}} & \textbf{\textcolor{red}{0.854}} & \textbf{\textcolor{red}{0.866}} & \textbf{0.900} & \textbf{0.855} & \textbf{0.866} & \textbf{0.850} & \textbf{0.803} & \textbf{0.815} \\
\hline
\end{tabular}%
\end{table*}

\begin{table}[t]
  \caption{Schema Matching (SM) Accuracy; Datasets \cite{liu2025magneto, koutras2021valentine}.}
  \label{tab:schema_matching}
  \vspace{-2ex}
  \centering
  \footnotesize
  \begin{tabular}{l|cc|cc|cc| cc}

    \hline
    \textbf{Dataset} & \multicolumn{2}{c|}{\textbf{ISResMat}} &
     \multicolumn{2}{c|}{\textbf{SimFlooding}} &
      \multicolumn{2}{c|}{\textbf{Magneto}} &
      \multicolumn{2}{c}{\textbf{\texttt{CONE}}} \\
     &
    R & MRR &
    R & MRR &
    R & MRR &
    R  & MRR \\
    \hline
    GDC & 0.24 & 0.30 & 0.24 & 0.29 & 0.50 & 0.87 & 0.65 & 0.85   \\
    Magellan  & 0.98 & 0.99 & 1.00 & 1.00 & 1.00 & 1.00 & 1.00 & 1.00 \\
    WikiData & 0.85 & 0.96 & 0.65 & 0.85 & 0.90 & 0.95 & 0.94 & 0.95  \\
    Open Data  & 0.60 & 0.78 & 0.45 & 0.75 & 0.75 & 0.95 & 0.88 & 0.90   \\
    ChEMBL  &  0.65 & 0.82 & 0.45 & 0.68 & 0.86 & 0.97 & 0.91 & 0.95  \\
    TPC-DI  & 0.82 & 0.93 & 0.68 & 0.82 & 0.85 & 0.97 & 0.92 & 0.94  \\
    \hline
  \end{tabular}
\end{table}


\subsubsection{Evaluation Process}
We construct sets of columns and tuples from tables in our datasets (Section \ref{sec:datasets}) that contain either strictly numerical values or mixed textual–numerical values. We use the method of \cite{kandibedala2025scalable} to identify attributes (e.g., hemoglobin, tumor size, sample size, blood loss, platelet count, etc.). Numerical values and their associated unit symbols are parsed using regular expressions and techniques from \cite{ptype-semantics}. To handle inconsistent unit representations (e.g., \textit{ml} vs.\ \textit{mL}), we apply \textit{unit canonicalization} \cite{ptype-semantics}, which normalizes all variants to a single canonical form. For entries with \textit{missing units}, we infer the most likely unit from the surrounding column or tuple context \cite{ptype-semantics}. When all entries in a column or tuple lack unit information, we assign a zero-padded vector for the unit component in the composite embedding. Following this pre-processing step, we compute the composite embeddings (Algorithm \ref{alg:composite} in Section \ref{sec:composite_embed}) for all columns and tuples in our dataset.

To enable scalable vector retrieval, we integrate the vector database system Meta Faiss \cite{douze2024faiss}, which indexes \texttt{CONE} embeddings using flat indexing via IndexFlatIP. We L2-normalize all embeddings prior to indexing, such that the inner-product search corresponds to cosine similarity. This method encodes vectors into fixed-size representations stored in a contiguous array, supporting high-throughput similarity search. Querying a column against 200K indexed vectors requires only 8.57\,ms to retrieve the top-10 most similar columns, demonstrating both scalability and low latency.

We manually curated the ground-truth matches with assistance from five volunteers and two independent domain experts; cases with inter-expert disagreement were excluded from the final evaluation set. The resulting benchmark contains 200 columns and 200 tuples with numerical values as references. For each reference column or tuple, we retrieve and rank the top-$K$ ($K=10$) candidate matches based on the \textit{cosine similarity} of their embedding vectors, sorted in descending order. The quality of these rankings is evaluated using Recall@$10$ \cite{kim2022exploiting}, Mean Average Precision (MAP@$10$) \cite{MAP}, and Mean Reciprocal Rank (MRR@$10$) \cite{MRR}. Recall@$K$ measures the proportion of relevant results retrieved within the top-$K$ candidates, while MAP and MRR capture both the precision and the ranking quality across multiple queries. All metrics are computed on the sorted list of clustered columns/tuples (ranked by cosine similarity in descending order) and averaged across all reference inputs.


We use cosine similarity, which measures the angle between vectors, to evaluate distance between embeddings of tuples or columns. Cosine similarity is a stable and widely used metric in embedding-based solutions for clustering and classification \cite{schutze2008introduction, rahutomo2012semantic, thongtan2019sentiment, ristanti2019cosine, li2013distance, kandibedala2025scalable} because it is not sensitive to the size of rows or columns in a table, unlike other methods \cite{ristanti2019cosine}. Relying solely on vector magnitudes, without considering angular distance, can cause rows or columns with similar content to appear significantly different in the embedding space. Alternative similarity metrics, such as Euclidean distance \cite{euclidean}, measure the absolute distance between two vectors in a vector space and are sensitive to their magnitudes, i.e. two vectors that are semantically similar but differ in scale might appear far apart. Jaccard similarity \cite{niwattanakul2013using} is designed for comparing sets and quantifies similarity based on the proportion of overlapping elements, which is different from semantic similarity measures used in vector spaces.

\begin{table*}[t]
  \caption{Ablation Study on Column Matching (CM) and Tuple Matching (TM): Average Recall@10 (R), MAP@10 and MRR@10\textsuperscript{\ref{fn:commonfootnote}}.}
  \label{tab:combined_ablation}
  \centering
  \footnotesize
  \begin{tabular}{l|l|ccc|ccc|ccc|ccc|ccc}
    \hline
    & & \multicolumn{3}{c|}{\textbf{\texttt{CONE}$_1$}} & 
      \multicolumn{3}{c|}{\textbf{\texttt{CONE}$_2$}} & 
      \multicolumn{3}{c|}{\textbf{\texttt{CONE}$_3$}} & 
      \multicolumn{3}{c|}{\textbf{\texttt{CONE}$_4$}} & 
      \multicolumn{3}{c}{\textbf{\texttt{CONE}}} \\
    \textbf{Dataset} & \textbf{Task} & R & MAP & MRR & R & MAP & MRR & R & MAP & MRR & R & MAP & MRR & R & MAP & MRR\\
    \hline
    \multirow{2}{*}{CancerKG} & CM & 0.800 & 0.789 & 0.808 & 0.766 & 0.764 & 0.772 & 0.790 & 0.773 & 0.786 & 0.797 & 0.782 & 0.790 & \textbf{0.833} & \textbf{0.817} & \textbf{0.831} \\
                              & TM & 0.833 & 0.800 & 0.804 & \textcolor{red}{0.700} & 0.783 & 0.791 & 0.820 & 0.790 & 0.802 & 0.830 & 0.796 & \textcolor{red}{0.807} & \ \textbf{\textcolor{red}{0.867}} & \textbf{0.844} & \textbf{\textcolor{red}{0.853}} \\
    \hline
    \multirow{2}{*}{CovidKG}  & CM & 0.767 & 0.752 & 0.770 & 0.733 & 0.732 & 0.746 & 0.760 & 0.748 & 0.771 & 0.765 & 0.754 & 0.768 & \textbf{0.800} & \textbf{0.782} & \textbf{0.798} \\
                              & TM & \textcolor{red}{0.700} & 0.724 & \textcolor{red}{0.736} & 0.700 & \textcolor{red}{0.658} & \textcolor{red}{0.688} & \textcolor{red}{0.720} & \textcolor{red}{0.705} & \textcolor{red}{0.725} & \textcolor{red}{0.752} & \textcolor{red}{0.723} & 0.755 & \textbf{\textcolor{red}{0.800}} & \textbf{\textcolor{red}{0.781}} & \textbf{\textcolor{red}{0.792}} \\
    \hline
    \multirow{2}{*}{WebTable} & CM & 0.900 & 0.806 & 0.814 & 0.900 & 0.801 & 0.802 & 0.920 & 0.818 & 0.832 & 0.920 & 0.821 & 0.835 & \textbf{0.950} & \textbf{0.842} & \textbf{0.858} \\
                              & TM & 0.880 & 0.820 & 0.830 & 0.870 & 0.821 & 0.830 & 0.876 & 0.821 & 0.831 & 0.878 & 0.823 & 0.832 & \textbf{0.900} & \textbf{0.854} & \textbf{0.866} \\
    \hline
    \multirow{2}{*}{CIUS}     & CM & \textcolor{red}{0.800} & \textcolor{red}{0.790} & 0.829 & 0.850 & 0.798 & 0.808 & 0.875 & 0.821 & 0.831 & 0.882 & 0.825 & 0.834 & \textbf{\textcolor{red}{0.900}} & \textbf{\textcolor{red}{0.851}} & \textbf{0.864} \\
                              & TM & 0.850 & 0.800 & \textcolor{red}{0.810} & 0.837 & 0.782 & 0.790 & 0.885 & 0.826 & 0.836 & 0.880 & 0.823 & 0.833 & \textbf{0.900} & \textbf{0.855} & \textbf{\textcolor{red}{0.866}} \\
    \hline
    \multirow{2}{*}{SAUS}     & CM & 0.850 & 0.815 & 0.812 & 0.850 & 0.801 & 0.800 & 0.884 & 0.826 & 0.835 & 0.880 & 0.824 & 0.833 & \textbf{0.900} & \textbf{0.849} & \textbf{0.861} \\
                              & TM & 0.800 & 0.784 & 0.793 & 0.781 & 0.753 & 0.763 & 0.830 & 0.794 & 0.792 & 0.830 & 0.786 & 0.797 & \textbf{0.850} & \textbf{0.803} & \textbf{0.815} \\
    \hline
  \end{tabular}
\end{table*}

\subsubsection{Results}
Table \ref{tab:baselien_comparison} illustrates the experimental results comparing \texttt{CONE} to the SOTA models with the best results highlighted in bold, and the delta between the worst and best scores is shown in red. We can see in Figure \ref{fig:angle1} that with the new composite embedding structure in \texttt{CONE}, the similarity score between {\em Age} and {\em Follow-up(month)} from Figure \ref{fig:num_table} decreased from 0.99 (using BioBERT embeddings) to 0.83 (using \texttt{CONE} embeddings, see the orange and green bars in Figure~\ref{fig:angle1}) which is significant. With BioBERT embeddings, the distance between the attributes was too small, making them nearly indistinguishable (Figure \ref{fig:angle1}). In contrast, the new distance calculated using \texttt{CONE} embeddings (0.83) allows to separate them, as expected, as these attributes are semantically different.

\begin{figure}[htbp]
 \centering
  \includegraphics[width=1.0\columnwidth,trim=0 10 0 20,clip] {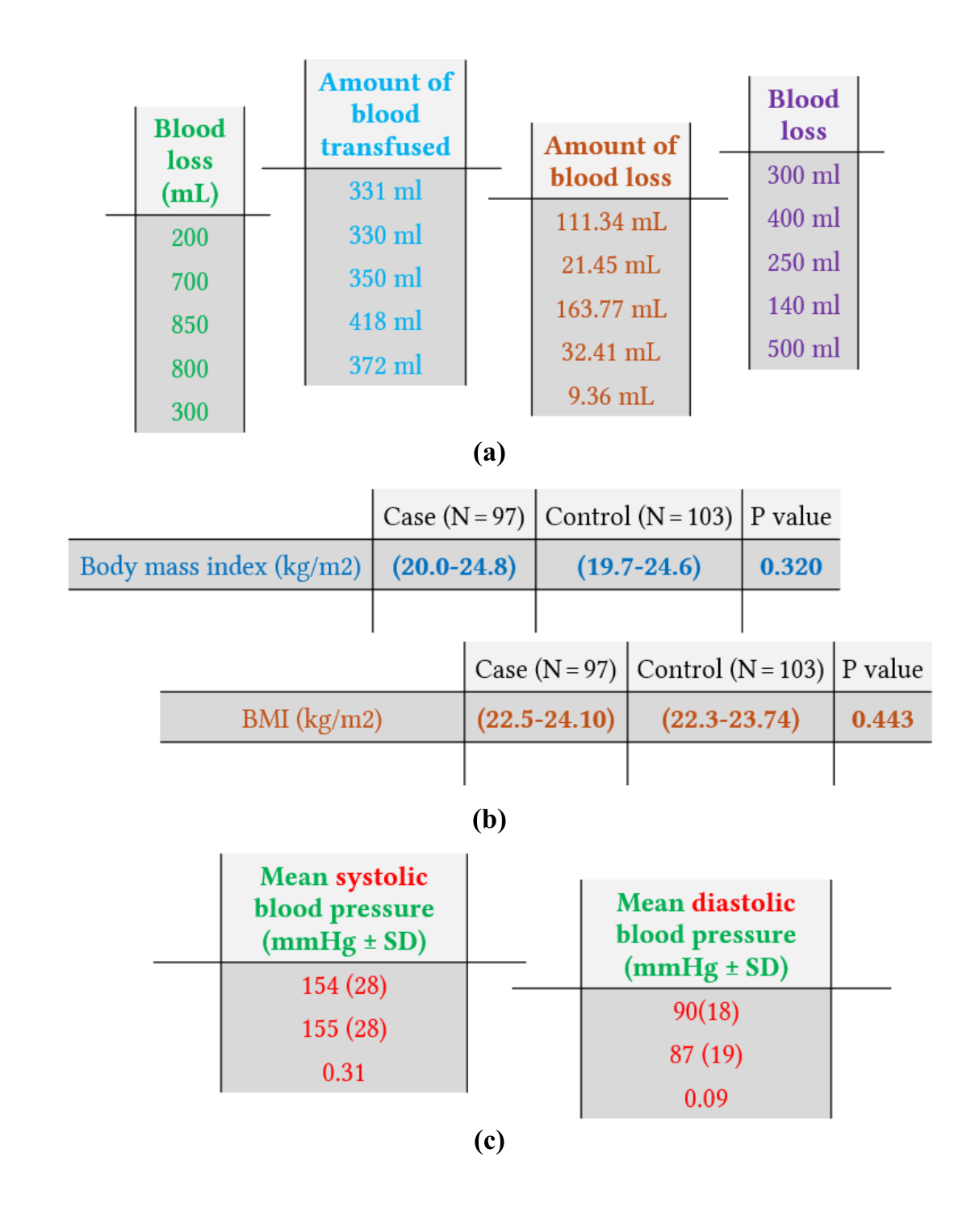}
  \captionsetup{skip=4pt} 
 \caption{Examples of \texttt{CONE} Composite Embeddings Correctly Matching Complex Semantically Similar (a) Columns; (b) Tuples; and (c) A Challenging Case of Column Matching.}
 \vspace{-2ex}  
 \label{fig:mered_example}
\end{figure}

It is important to note that our composite embeddings are resistant to \textit{attribute naming heterogeneity}. They understand and correctly match the same columns/tuples with different attribute names (e.g., `\textit{Blood Loss (mL)}', `\textit{Amount of blood transfused}', etc. in Figure \ref{fig:mered_example}(a)). Also, we are able to match column/tuple even if one of the pairs has attribute name in the abbreviated format(e.g., `\textit{BMI (kg/m2}') and `\textit{Body mass index (kg/m2)}' in Figure \ref{fig:mered_example}(b), however, the baseline method was unable to identify this match). 
Similarly, for a column \textit{cholesterol\_ldl mg/dL}, a naive text embedding of the attribute name and unit retrieves several unrelated attributes (e.g.,  \textit{bilirubin\_total}, \textit{hemoglobin}, \textit{albumin}, \textit{free\_t4}, etc.), while \texttt{CONE} avoids such spurious matches and correctly retrieves related attributes (e.g. \textit{cholesterol\_hdl}).
In summary, attribute naming variation is naturally handled by contextual embeddings that capture semantic similarity in the attribute space based on the context.

\noindent\textbf{Column Matching (CM):} \texttt{CONE} outperforms all the baselines across all datasets in the column matching task (Table \ref{tab:baselien_comparison}). It achieves the highest Recall@10, MAP@10, and MRR@10 scores, with a notable Recall@10 improvement of 25\% over NumNet on Webtables. \texttt{CONE} achieves the highest Recall@10 of 95\% on Webtables. 

\noindent\textbf{Tuple Matching (TM):}  In Table \ref{tab:baselien_comparison}, we can see that \texttt{CONE} achieves the best overall performance across all metrics and datasets for tuple matching. \texttt{CONE}  outperforms NumNet with a significant Recall@10 delta of 17.7\% on Webtables.

The consistent MAP and MRR gains further confirm that \texttt{CONE} not only retrieves the correct columns/tuples but also ranks them more accurately within the top results.

\noindent\textbf{Schema Matching (SM):} We conduct additional SM experiments on six datasets from  \cite{liu2025magneto}. The results are shown in Table \ref{tab:schema_matching}. \texttt{CONE} achieves comparable or better Recall (R) than all baselines, demonstrating the effectiveness of its explicit encoding of attribute, numerical value, and unit semantics. Although Magneto attains slightly higher MRR due to its LLM-based re-ranking stage, our approach achieves comparable MRR without LLM calls, making it a more cost-efficient alternative.

\vspace{-1.5ex}
\subsection{Ablation Study}
We evaluated  four variants of \texttt{CONE}: \texttt{CONE}$_1$, where we removed the numerical module (i.e., the numerical value embedding component, dotted box in Figure \ref{fig:architecture}); \texttt{CONE}$_2$, where we removed the composite embedding structure (Figure \ref{fig:comp}(a)), eliminating the joint embedding of attribute, unit, and value; \texttt{CONE}$_3$, where we removed the unit component from the composite embedding; and \texttt{CONE}$_4$, where we removed the encoding for numerical Ranges and Gaussians. Table~\ref{tab:combined_ablation} illustrates our observation that removal of any component leads to drop in Recall, MAP, and MRR across all datasets for both CM and TM tasks. \texttt{CONE}$_1$ demonstrates the Recall drop of 10\% for CM on CIUS and TM on CovidKG. \texttt{CONE}$_2$ demonstrates the most significant performance degradation in TM, with 16.7\% Recall decrease on CancerKG. \texttt{CONE}$_3$ reduces Recall by up to 8\% on TM for CovidKG. Finally, \texttt{CONE}$_4$ leads to Recall reduction of up to 4.8\% on TM for CovidKG. We conclude that removing either numerical module or composite embedding structure significantly hurts performance as evidenced by these ablation study results.

\subsection{Challenging Cases}
Figure \ref{fig:mered_example}(c) presents an example of a challenging case (CovidKG dataset). We observed cases where columns with highly similar attribute names (e.g., `\textit{Mean Systolic Blood Pressure}' vs. `\textit{Mean Diastolic Blood Pressure}' with $\approx$ 75\% lexical overlap highlighted in green) and identical units (e.g., \textit{mmHg $\pm$ SD}) are embedded close together despite having significant different values. The key reason is {\em very rare} occurrence of columns with such unit in the dataset (e.g., both \textit{systolic} and \textit{diastolic} attribute with this unit occur only once in the entire CancerKG dataset). Inaccuracy of the embedding vectors for such rare entities is a common drawback of any architecture, not only ours, because some reasonable number of occurrences of an entity is needed for the model to collect different contexts required for accurate embedding vector formation. The smaller the number of occurrences, the more inaccurate the embedding vector representation will get in any of existing SOTA architectures. 

\vspace{-2ex}
\section{Related Work} \label{sec:related_work}
Traditional word embeddings techniques \cite{mikolov2013distributed, lund1996producing,pennington2014glove, BERT} have shown strong performance in language modeling but inadequately represent numerals. They treat numbers as regular words, ignoring their magnitude, unit, and context \cite{naik2019exploring}, and numerals are often underrepresented in training corpora—only 3.79\% of GloVe’s 400,000 embeddings correspond to numeric tokens \cite{pennington2014glove}. 

Several studies propose numeric-aware models. \cite{spithourakis2018numeracy} model numerals separately using continuous density functions, improving perplexity and prediction accuracy. \cite{jiang2020learning} address out-of-vocabulary issue for numerals by inducing prototype embeddings via Gaussian mixtures or self-organizing map, integrated into word embedding training. DICE (Deterministic Independent-of-Corpus Embeddings) \cite{sundararaman2020methods} handcrafts embeddings to preserve numeric numeric distances using cosine and Euclidean measures, aiding tasks like comparison and addition. Our approach builds on these foundations by encoding complex numerical data by decomposing them into their attributes, the values, and associated units, thereby allowing for a more fine-grained semantic distinction. We incorporate DICE as one component of our composite embedding but extend it by incorporating attribute and unit information, to address polysemy (e.g., \textit{5 years} vs. \textit{5 miles}) and represent ranges and Gaussians. 

Earlier efforts on numeric reasoning in Transformer-based models \cite{ashish2017attention} include NumNet \cite{ran2019numnet}, which represents numbers in graphs but lacks contextual and type awareness. AeNER \cite{yarullin2023numerical} proposed attention-enhanced numerical embeddings for numerical reasoning over text and tables. It enriches BERT with numerically-aware embeddings and attention, but does not distinguish units or support numerical ranges. NumBERT \cite{zhang2020language} and NumGPT \cite{jin2021numgpt} use scientific notation and prototype encodings for numbers, but their effectiveness in capturing the semantics of complex structured data is not well explored. GenBERT \cite{genbert} is a QA model with pretrained BERT serving as both its encoder and decoder. It uses the decimal notation and digit-by-digit tokenization of numbers. NC-BERT \cite{kim2022exploiting} extends GenBERT by adding a numerical-contextual masking layer on top of BERT to emphasize number-related contextual information. 

Recent works, such as xVal \cite{golkar2023xval} and NTL \cite{zausinger2025regress}, highlight the need for explicit numeric inductive biases in language models. NTL proposes a regression-like loss function for numeric decoding as an alternative to the standard cross-entropy loss. MMD \cite{alberts2024interleaving} encodes numbers similar to xVal, but introduces a routing layer to explicitly separate text and number representations. However, unlike us, these methods do not explicitly incorporate attributes or units, limiting their applicability to structured numeric data.
FoNE \cite{ zhou2025fone} introduces an alternative numerical embedding strategy for LLMs by encoding numbers with cosine–sine Fourier features, but it only captures syntactic structure and ignores semantic aspects such as attribute, value, and unit. NumeroLogic \cite{schwartz2024numerologic} prefixes numbers with digit counts, giving LLM place-value awareness and inducing a chain-of-thought step during number generation. MultivariateGPT \cite{loza2025multivariategpt} improves how models predict numbers and categories together, while our work focuses on improving how numbers are embedded. \cite{sivakumar2025digit} uses mathematical priors to aggregate digit embeddings to improve scalar magnitude representations; in contrast, \texttt{CONE} encodes not only scalars, but also numerical ranges and gaussians.

Other approaches focus on numeric features in tabular data. \cite{gorishniy2022embeddings} propose encodings such as Piecewise Linear Encoding for Multilayer Perceptron and Transformer models, but their methods are designed for supervised settings.
In contrast, our approach learns representations in an unsupervised manner to capture the structure of numeric tables. 
TAPAS~\cite{herzig2020tapas} introduces rank/row/column embeddings for tables, and similar structural encodings are used by TUTA~\cite{TUTA}, TURL~\cite{TURL}, 
TaBERT~\cite{TaBERT}, and TABBIE~\cite{TABBIE}. However, these models blindly treat numbers and words the same when deriving 
embedding vectors, whereas we treat them differently in our approach.
CARTE \cite{kim2024carte} introduces a graph-based, pre-trained model for tabular learning, but does not explicitly model numerical semantics.
Schema matching systems such as Rondo \cite{MelnikRB03} and model management frameworks \cite{mm, mm2} typically rely on attribute names and data values to infer matches, but do not explicitly model units or distinguish numerical magnitudes and ranges. 
Valentine \cite{koutras2021valentine} provides a comprehensive benchmark for evaluating schema matching methods, while Magneto \cite{liu2025magneto} introduces an LLM-assisted matching system based on embedding retrieval and re-ranking; however, both do not encode numerical representations. In contrast, \texttt{CONE} addresses this gap by learning numerical embeddings that explicitly encode magnitude, units, and attribute context, enabling more effective column, tuple, or schema matching.

To the best of our knowledge, none of the existing embedding methods can handle numeral polysemy. For example, the numeral `\textit{2019}' may denote either a year or an ordinary number, and numerical expressions like `\textit{7 lb}' and `\textit{7 miles}' have distinct semantic interpretations. Our method addresses this challenge by explicitly concatenating each individual components that define semantic of numerical in structured data i.e the attribute name to categorize particular entity, the numerical value and the unit. Furthermore, we introduce specialized embeddings for representing numerical ranges and gaussians.

\vspace{-2ex}  
\section{Conclusion}\label{sec:conclusion}
We introduced \texttt{CONE}, a context-aware embedding model that preserves the semantics of numerical values, ranges, and gaussians by jointly encoding values, units, and attribute names using a composite embedding vector structure. To achieve this, we design a novel composite embedding construction algorithm that uses pre-computed \texttt{CONE} embeddings. In numerical reasoning evaluation, \texttt{CONE} achieves an F1 score of 87.28\% on DROP, outperforming SOTA baselines by up to 9.37\% F1. Additionally, experiments on several large-scale structured datasets demonstrate that \texttt{CONE} surpasses the major SOTA baselines on two popular downstream tasks, achieving a significant Recall@10 improvement of up to 25\%, thereby demonstrating its effectiveness in encoding structured numerical data.


\bibliography{bibfile}
\bibliographystyle{ACM-Reference-Format}
\end{document}